\pdfoutput=1
\documentclass[11pt]{article}

\usepackage[]{acl}

\usepackage{times}
\usepackage{latexsym}

\usepackage[T1]{fontenc}
\usepackage[utf8]{inputenc}

\usepackage{microtype}

\usepackage{inconsolata}

\usepackage{graphicx}

\usepackage{booktabs}
\usepackage{multirow}
\usepackage{subcaption}
\usepackage{rotating}
\usepackage{amssymb}% http://ctan.org/pkg/amssymb
\usepackage{pifont}% http://ctan.org/pkg/pifont

\usepackage{colortbl}       % for coloured rows in tables

\newcommand{\cmark}{\ding{51}}%
\newcommand{\xmark}{\ding{55}}%

\usepackage{minitoc}

\title{\emph{Around the World in 24 Hours}\\ Probing LLM Knowledge of Time and Place}
\author{
  \textbf{Carolin Holtermann\textsuperscript{1}},
  \textbf{Paul Röttger\textsuperscript{2}},
  \textbf{Anne Lauscher\textsuperscript{1}}
\\
  \textsuperscript{\rm 1}University of Hamburg, Germany\\
    \textsuperscript{\rm 2}University of Bocconi, Italy\\
\\
  \texttt{carolin.holtermann@uni-hamburg.de}
}

\begin{document}
\doparttoc % Tell to minitoc to generate a toc for the parts
\faketableofcontents % Run a fake tableofcontents command for the partocs

%\part{} % Start the document part
%\parttoc % Insert the document TOC

\maketitle

\begin{abstract}
%Towards the ultimate development of a world model, the understanding of time and space is essential to accurately represent the physical world and provide contextualized responses to a variety of real-world scenarios. Prior work has tested temporal and geographical knowledge in language models, but only in simple settings and separately from one another.

Reasoning over time and space is essential for understanding our world. However, the abilities of language models in this area are largely unexplored as previous work has tested their abilities for logical reasoning in terms of time and space in isolation or only in simple or artificial environments. %However, previous work has tested language models' abilities for geographical and temporal reasoning in isolation, or in simple or artificial settings only.
%Although previous work has tested the abilities of temporal reasoning and geographical understanding of the world of large language models in simple settings and independently, the integrated analysis of how these models connect time and space remains unexplored.
In this paper, we present the first evaluation of the ability of language models to \textit{jointly} reason over time and space.
%This study introduces a comprehensive examination of language models' ability to combine gathered geospatial and temporal knowledge despite only their factual recall.
To enable our analysis, we create \textsc{GeoTemp}, a dataset of 320k prompts covering 289 cities in 217 countries and 37 time zones.
%and varying levels of query complexity that serves as the first benchmark evaluating the interception of temporal and geographical knowledge.
Using \textsc{GeoTemp}, we {\color{black} evaluate eight open chat models from three model families for different combinations of temporal and geographic knowledge.}
We find that most models perform well on reasoning tasks involving only temporal knowledge and that overall performance improves with scale.
However, performance remains poor in tasks that require connecting temporal and geographical information. 
We do not find clear correlations of performance with specific geographic regions. Instead, we find a significant performance increase for location names with low model perplexity, {\color{black} suggesting their repeated occurrence during model training}. We further demonstrate that model performance is heavily influenced by prompt formulation -- a direct injection of geographical knowledge leads to performance gains, whereas, surprisingly, techniques like chain-of-thought prompting decrease performance on simpler tasks.\footnote{We release all data and code at \url{https://github.com/UhhDS/GeoTemp}.}

%Previous work found biases in the performance of language models relating to that cultural biases 
%How much do models know about the world
%geotemporal knowledge
%Current models struggle in particular with non-western countries
%we find improvements in scale
%we also find that how much world knowledge can be abgerufen hängt vom prompt ab
%How good the models perform is dependent on their perplexity
%we evaluate on city level(e.g. some countries are large and have multiple time zones)
%we present a novel carefully curated dataset including ... countries, ... time zones, diverse difficulty levels of size ... 

\end{abstract}

\section{Introduction}
\begin{quote}
    \emph{``Stop worrying about the world ending today. It's already tomorrow in Australia.''}
— Charles M. Schulz
\end{quote}

%\begin{quote}
%"Time is relative."
%— Albert Einstein
%\end{quote}

\begin{figure}
    \centering
    \includegraphics[width=0.70\linewidth]{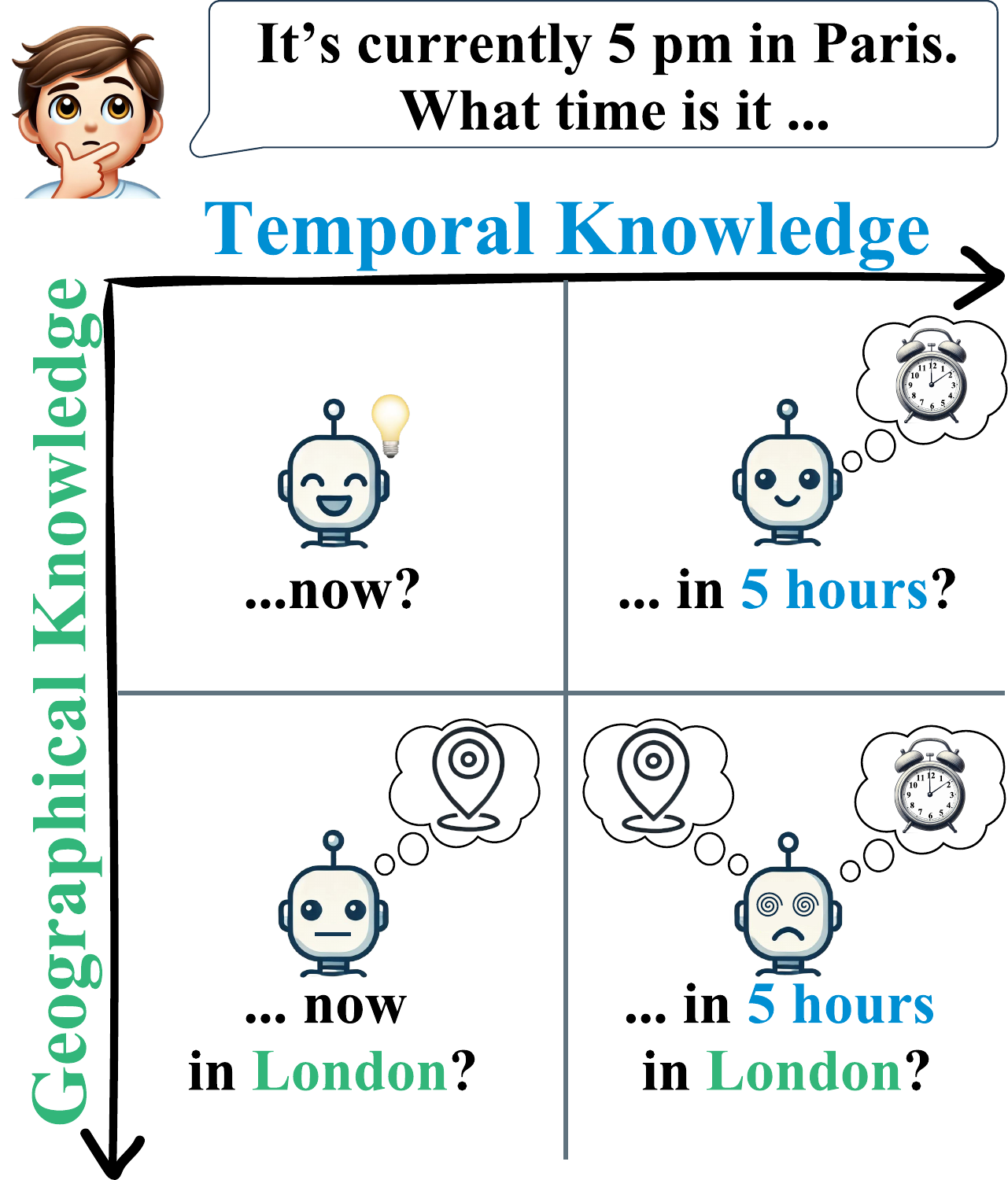}
    \caption{\textbf{Combinations of temporal and geographic knowledge we test with \textsc{GeoTemp}.}
    Models struggle most with jointly reasoning over space and time.}
    \label{fig:combinations}
\end{figure}

% Instructions / ideas for the intro :)
% Start with something general -- actually, I loved the quote also that we have now put into the other paper :D .. well
% the first sentence is really for everyone, also your mom
% The second sentence can be more specific an situate this in the NLP realm
% Important is to outline why this is important
% Then, some RW
% Oh no! there is a research gap
% We address this research gap
% How? + findings....
% Exact contributions (and maybe structure of the paper)

%\noindent As human civilization has progressed, systems and conventions have been devised to make life more organized and communal. This development includes, among other things, a globally standardized time and calendar system, the creation of time zones based on geographical location, and the introduction of daylight saving time to synchronize activities across the world \cite{evolution_of_time}. These conventions have become ingrained in our fundamental world knowledge, shaping our understanding of the relationship between time and space.
%Human civilization has developed various systems and conventions to organize life and foster a sense of community. These include a globally standardized time and calendar system, the division of time zones based on geographic location, and the implementation of daylight saving time to align activities across different regions \cite{evolution_of_time}.

\noindent Human civilization has developed systems and conventions to organize life and foster community, such as a standardized global time and calendar system, time zones based on geographic location, and daylight saving time to align activities across different regions \cite{evolution_of_time}. {\color{black} These conventions, deeply rooted in our understanding of the world, play a crucial role in decision-making, especially in today's globalized working environment. As technology advances, large language models (LLMs) will increasingly be used to simplify and optimize tasks \cite{eloundou2023gptsgptsearlylook}, for example in the logistical planning of cross-border shipping or the optimal planning of business trips. To reliably perform such tasks, they must be able to incorporate temporal and geographical information into their reasoning process. }%These conventions are now deeply embedded in our understanding of the world, influencing how we perceive the relationship between time and space.
%Large language models (LLMs), which are now used by millions of people across the world \citep{reuters2024openai}, must also have a basic understanding of these \textit{geotemporal} concepts to be useful for everyday tasks such as determining the best time to contact a colleague in Australia or plan travel.
Prior work has evaluated the temporal reasoning ability of LLMs ~\cite[e.g.][]{templama, fatemi2024testtimebenchmarkevaluating} as well as their spatial understanding of the world ~\cite[e.g.][]{gurnee2024languagemodelsrepresentspace, worldbench}.
However, little work has been done on evaluating LLM ability to reason \textit{jointly} over space and time, outside of simple and artificial settings such as multiple-choice questions~\cite{wang-zhao-2024-tram} or synthetic data ~\cite{fatemi2024testtimebenchmarkevaluating}.

%To be truly useful in many everyday tasks – e.g., determining the best time to contact a colleague in Australia – language models must also have a basic understanding of these concepts. Accordingly, previous research has thoroughly examined the abilities of large language models (LLMs) in temporal reasoning~\cite[e.g.,][]{templama, fatemi2024testtimebenchmarkevaluating}, and has explored their understanding of the global structure of the world~\cite[e.g.,][]{gurnee2024languagemodelsrepresentspace, worldbench}. However, the research landscape on the connection of temporal and geographical knowledge remains scarce: while initial works on testing geotemporal reasoning in simple and artificial settings (i.e., on easy multiple choice questions~\cite{wang-zhao-2024-tram}, and on synthetic data~\cite{fatemi2024testtimebenchmarkevaluating}) exist, to-date there is no single work on the interplay of geographical and temporal knowledge in realistic settings. 

In this paper, we address this gap by analyzing the ability of LLMs to combine geographical and temporal knowledge in reasoning tasks, using global time zones as a testbed. To this end, we present \textsc{GeoTemp}, which contains 320k question prompts covering 289 locations in 37 time zones.
\textsc{GeoTemp} is explicitly designed to evaluate LLM across different levels of temporal and geographical reasoning (Figure~\ref{fig:combinations}). Using \textsc{GeoTemp} we answer the following research questions:

\noindent \textbf{1) How well do LLMs reason about time, place, and their combination?}
We test eight state-of-the-art chat-optimized LLMs on four different tasks requiring different levels of temporal and/or geographical knowledge (\S\ref{sec:reason}). We find that most models are able to perform simple calculations of time, but fail as soon as they have to perform additional geographical reasoning. Although we see improvements with model scale, the best-performing model is only 25.4\% accurate on questions requiring both temporal and geographical knowledge. %able to answer 25.4\% of the questions correctly for the most complex prompt task.

\noindent \textbf{2) What factors influence model performance on geotemporal tasks?}
In contrast to prior work on geographic knowledge~\cite{worldbench}, we do not find clear performance disparities across geographic regions  (\S\ref{sec:factors}). However, we do observe a significant correlation between model performance and location name perplexities. Direct probing of time zones reveals that models generally possess knowledge of individual geographical and temporal facts, but are unable to combine the two effectively, which limits performance on geotemporal tasks.

\noindent \textbf{3) Can model performance be improved by injecting geotemporal knowledge?}
We provide models with location-specific time zones to guide their reasoning (\S\ref{sec:knowledge_inj}), and find significant performance improvements.
However, surprisingly, these improvements are mostly for locations for which, in our prior probing, we found the models to know the time zone of already.
This suggests that, while basic geotemporal facts may be already present in the models, this knowledge requires activation to solve more complex geotemporal tasks.

\section{Related Work}
\paragraph{Testing temporal knowledge}
Several studies benchmark the temporal knowledge of LLMs. The TimeBank corpus~\cite{timebank} focuses on annotating news articles for temporal expressions, events, and their links. Recent work expands to analyze temporal knowledge beyond event extraction, with benchmarks like McTaco \cite{zhou-etal-2019-going}, TempLama \cite{templama}, or SituatedQA \cite{zhang-choi-2021-situatedqa} probing models' factual knowledge of time-varying. Other studies explore their temporal reasoning capabilities, including relations between temporal expressions \cite{thukral2021probinglanguagemodelsunderstanding}, event ordering \cite{tan-etal-2023-towards, wei-etal-2023-menatqa, xiong-etal-2024-large}, and duration \cite{vashishtha-etal-2020-temporal}, and more complex tasks like multi-hop reasoning \cite{tan2024robusttemporalreasoninglarge} or video reasoning \cite{videoinsta}. Recent work also examines the temporal knowledge alignment of LLMs to given timestamps \cite{realtimeqa, streamingqa, zhao2024setclocktemporalalignment}. Finally, \citet{nylund-etal-2024-time} and \citet{gurnee2024languagemodelsrepresentspace} analyze temporal representations within models through probing and methods from task arithmetic. % to manipulate time vectors, aiming to enhance models' temporal reasoning capabilities.

%CogCompTime \cite{ning-etal-2018-cogcomptime} a package to extract temporal expression from text

\paragraph{Testing geographical knowledge}
Previous work on geographic knowledge in LLMs focuses on the prediction of geolocations \cite{godey-etal-2024-scaling}, their internal model representation \cite{gurnee2024languagemodelsrepresentspace}, and how this knowledge can be adapted \cite{lietard-etal-2021-language}. Other works extend these efforts by evaluating the ability of LLMs to recall information about different countries and examine model-intrinsic and location-specific biases \cite{worldbench, manvi2024geollmextractinggeospatialknowledge, manvi2024largelanguagemodelsgeographically}.

%determine people's location based on their posted tweets \cite{xiao-etal-2024-analyzing, xiao-etal-2023-context}

\paragraph{Testing geotemporal knowledge}
Two recent works partially address geotemporal reasoning but provide limited insight into the ability of models to combine geographical and temporal knowledge.
1)~The TRAM corpus \cite{wang-zhao-2024-tram} evaluates temporal reasoning, including time zone calculations as a subtask. However, it focuses on model performance and does not provide a deeper analysis of the model's ability to combine temporal and geographical knowledge. Moreover, its multiple-choice format does not allow open-ended responses and thus strongly guides the model in its response. 2)~The TOT benchmark \cite{fatemi2024testtimebenchmarkevaluating} is a fully synthetic data set that tests temporal reasoning without including real-world location information. It evaluates time zone calculations, but based on the provision of time zones, not geographic locations.

\section{\textsc{GeoTemp}}
{\color{black} We introduce \textsc{GeoTemp}, the first probing dataset specifically designed to evaluate the geotemporal knowledge of LLMs. By carefully constructing synthetic prompts that progressively require more complex geotemporal reasoning, \textsc{GeoTemp} allows us to isolate failure points while minimizing the risk of using exact prompts seen during training. However, grounded in real-world LLM usage, \textsc{GeoTemp} combines temporal and geographical knowledge, using global time zones as its basis.}

%With \textsc{GeoTemp}, we present the first benchmark dataset specifically designed to assess the geotemporal knowledge of LLMs across time, place, and combinations thereof. Global time zones provide an ideal foundation for this, allowing us to evaluate the ability of models to make use of both temporal and geographical knowledge in a context grounded in real-world LLM usage.

%%%%%%%%%%%%%%%%%%%%%%%%%%%%%%%%%%%%%%%%%%%%%%%
\subsection{Dataset Creation}
We create \textsc{GeoTemp} in three steps: 1)~We collect time zones along with cities located in them; 2)~We craft four task templates representing varying combinations of geotemporal knowledge, coupling each of these tasks with one or two of our locations, depending on the task template; 3)~We compose the final dataset by sampling representative task-location combinations from above, and embedding them into the task templates.

\paragraph{1) Collecting time zones and locations}

We use the Olson Time Zone Database (OTZD) to compile a diverse selection of locations across countries and time zones.\footnote{\url{https://www.iana.org/time-zones}}
The OTZD contains time zone information for 596 representative locations worldwide, including rules for daylight saving time. % and historical changes to time zone boundaries by political authorities.
It is actively maintained by the non-profit organization ICANN, ensuring reliability and accuracy.\footnote{\url{https://www.icann.org/}}
% We access the OTZD using the Python libraries \texttt{pytz} and \texttt{datetime}.Time zone data is accessible through Python using the pytz package\footnote{\url{https://pypi.org/project/pytz/}}, in combination with the datetime package\footnote{\url{https://docs.python.org/3/library/datetime.html}}, facilitating a seamless integration for efficient transformation and evaluation of time data.

\setlength{\tabcolsep}{9pt}
\begin{table}[t]
    \centering
    \small
    \begin{tabular}{l l}
    \toprule
       \textbf{Task Name}  &  \textbf{Template}\\
       \midrule
       \rowcolor[HTML]{EFEFEF}
       \textsc{Verification} & %Today is \textit{\{Time\&Date\}} in \textit{\{City1\}}. 
       \emph{What time is it now in $l_1$?}  \\
       \textsc{TimeTime} & %Today is \textit{\{Time\&Date\}} in \textit{\{City1\}}. 
       \emph{What time is it in \textit{x} hours?} \\
       \rowcolor[HTML]{EFEFEF}
       \textsc{TimePlace} & %Today is \textit{\{Time\&Date\}} in \textit{\{City1\}}. 
       \emph{What time is it now in $l_2$?}  \\
       \textsc{TimeTimePlace} & %Today is \textit{\{Time\&Date\}} in \textit{\{City1\}}. 
       \emph{What time is it in \textit{x} hours in $l_2$?} \\
       \bottomrule
    \end{tabular}
    \caption{\textbf{Overview of tasks and their  templates} we use to construct test prompts in \textsc{GeoTemp}.
    We prepend to each template information about the current date, time, and location as follows: \emph{Today is \textit{\{Time\&Date\}} in \textit{\{l$_1$\}}}.}
    \label{tab:prompttypes}
\end{table}

The 596 timezone names we extract from the OTZD typically follow a naming convention of \textit{Area/City} (e.g.\ \textit{Europe/Madrid}).
To ensure that all entries in our dataset refer to precise locations, we manually exclude time zones that represent entire regions, such as \textit{Brazil/West}, as well as outdated time zones that are no longer in use. After applying these filters, we are left with a refined set of 460 distinct and precise locations. 
For each of these locations, we gather additional \textit{city}-level data, including population, latitude, and longitude, using the Opendatasoft API, to enable more comprehensive analysis later on.\footnote{\url{https://public.opendatasoft.com/explore/dataset/geonames-all-cities-with-a-population-500/api/?disjunctive.country}}

%Specifically, we retrieve \textit{country}-level data, including country population, income level, and regional classification, from the World Bank database.\footnote{\url{https://data.worldbank.org/}}

%An automatic extraction is carried out using the city name and the time zone identifier. This API was chosen in particular as it is an open service that provides data for cities with as few as 500 inhabitants.

%%%%%%%%%%%%%%%%%%%%%%%%%%%%%%%%%%%%%%%%%%%%%%%
\paragraph{2) Crafting task templates}

%Next, we use the compiled location and time zone information to construct the prompts for the final evaluation dataset. 
Next, we define four tasks requiring different combinations of geotemporal knowledge. We present these in Table \ref{tab:prompttypes}. %, each designed to evaluate the geotemporal knowledge of language models at progressively higher levels of complexity, as shown in Table \ref{tab:prompttypes}.

\textbf{\textsc{Verification}} provides the model with a date and time in a given location $l_1$, and asks for that same time again (e.g.\ \emph{Today is Monday, June 19th at 10:33 AM in 2023 in Novokuznetsk, Russia. What time is it now in Novokuznetsk, Russia?}).
This allows us to assess the model's basic comprehension of the task and its ability to respond with a correctly formatted time. %Furthermore, we can verify whether the model is using the provided information correctly, instead of relying on real-time data from the location in question.

\textbf{\textsc{TimeTime}} provides a date and time and a specific location $l_1$, then asks what the time will be in that same location in a specified number of hours (e.g.\ \emph{What time is it in 3 hours?}).
To accomplish this task, the model only needs to perform basic temporal reasoning, whereas geographical reasoning is irrelevant.
This template is adapted from prior work on LLM temporal knowledge and reasoning \cite{templama, tan2024robusttemporalreasoninglarge}.

\textbf{\textsc{TimePlace}} gives the model the current date and time at a specified location $l_1$ and asks for the current time at another given location $l_2$. To solve this task, the model requires knowledge of the time zones at each location and must determine the resulting time zone difference.

\textbf{\textsc{TimeTimePlace}} provides the model with a date and time at a specific location $l_1$, then asks the model for the time in another location $l_2$ but in a specific number of hours. Solving this requires the model to consider the time shift before or after the time zone change, in addition to \textsc{TimePlace}. This tests the model's ability to handle both temporal shifts and geographic knowledge simultaneously.

%%%%%%%%%%%%%%%%%%%%%%%%%%%%%%%%%%%%%%%%%%%%%%%
\paragraph{3) Composing the test set}
With \textsc{GeoTemp}, we want to cover a wide variety of locations while avoiding over-representing any particular combination of locations. To do this, we select a representative sample from the 460 locations obtained from OTZD, such that all 37 UTC time zones and 217 ISO country codes are included. This results in a refined selection of 289 locations. Next, we generate the Cartesian product of all selected locations $l_1 \times l_2$ to cover all possible combinations within the \textsc{TimePlace} and \textsc{TimeTimePlace} templates. Finally, we construct all evaluation prompts of \textsc{GeoTemp} by applying every task template of Table \ref{tab:prompttypes} to each $l_1 \times l_2$ location combinations. To avoid confusion with city names that exist in multiple countries (e.g.\ London in the US vs.\ London in the UK), we include the country information alongside the city name in the prompt. For each input prompt, we select the time and date at random within the year 2023, i.e. a recent year, to avoid outdated references and automatically calculate the target time and date which serves as the ground truth label for our evaluation algorithm. This results in a final set of 332,928 test prompts.

\begin{figure}[t]
    \centering
    \includegraphics[width=0.98\linewidth]{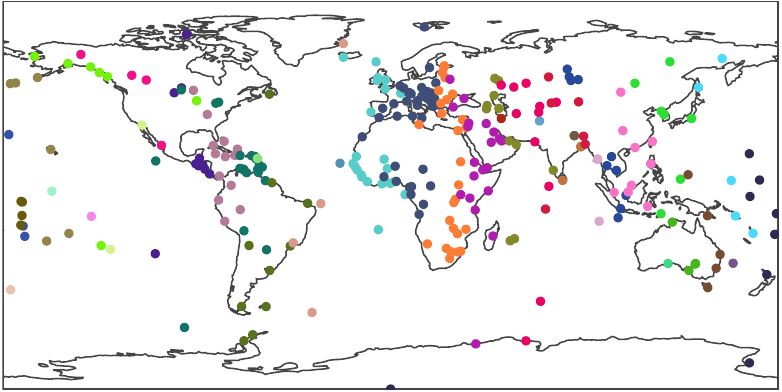}
    \caption{\textbf{Geographical distribution of all 289 locations in \textsc{GeoTemp}.} Color indicates UTC time zone.}
    \label{fig:distribution_locations}
\end{figure}
\begin{figure}[t]
    \centering
    \includegraphics[width=0.95\linewidth]{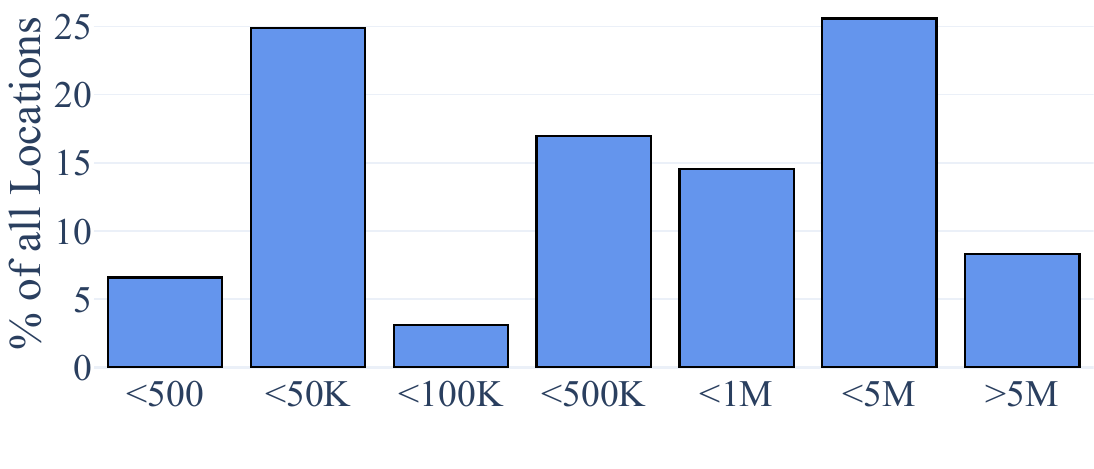}
    \caption{\textbf{Distribution of cities included in \textsc{GeoTemp} according to their number of inhabitants.} For each bin, we show the percentage of locations in our dataset that belong to this bin.}
    \label{fig:distribution_population}
\end{figure}

\subsection{Dataset Analysis}
\textsc{GeoTemp} covers a diverse set of locations and time zones across all continents, as shown in Figure \ref{fig:distribution_locations}, including Antarctica and several islands. \textsc{GeoTemp} also represents smaller and therefore often less well-known cities, such as ``Atikokan'' in Canada. The distribution of the number of inhabitants per city is shown in Figure \ref{fig:distribution_population}. More than 6\% of the locations in our dataset have $\leq$500 inhabitants, and 50\% of the cities have $\leq$500k inhabitants. For a detailed statistical analysis of the location information in \textsc{GeoTemp}, see Appendix \ref{app:dataset_dist}.

\section{Overall Experimental Setup}
\subsection{Models and Inference} \label{sec:experiments}
We test eight openly accessible chat-optimized LLMs spanning three model families and different model sizes on \textsc{GeoTemp}.
All models have shown competitive performance on the LMSyS Leaderboard.\footnote{\url{https://huggingface.co/spaces/lmsys/chatbot-arena-leaderboard}} Specifically, we evaluate Llama2-Chat \cite{touvron2023llama2openfoundation} in the model sizes 7B, 13B and 70B, Llama3-Instruct \cite{llama3modelcard} in both available model sizes (8B and 70B) and Qwen2-Instruct \cite{yang2024qwen2technicalreport} in the model sizes 1.5B, 7B and 72B. %As of July 2024, when we conducted our experiments, these represented the latest model versions.
We run all models on 1-4 Nvidia A6000 GPUs using the \texttt{simplegen} python library \cite{milanlp-2023-simple-generation}. In all experiments, we use a maximum length of 256 and a temperature of 0 to make model responses deterministic.

During inference, we run each prompt of \textsc{GeoTemp} using three instruction templates to analyze their effect on model performance. The first instruction type (\emph{neutral}) is shown in Table \ref{tab:prompttypes} and provides no additional information on how the model should answer the question. This leaves the model the ability to choose its own explanation methodology. The second type appends the suffix \emph{``Think step by step.''}, asking the model to apply \emph{chain-of-thought} (\emph{CoT}) reasoning. Lastly, the third type appends the suffix \emph{``Keep your answer short and just answer with the correct time and date''}, which encourages concise responses (\emph{short}).%\footnote{We publish all data and code at [REDACTED].}

\subsection{Evaluation Protocol: Regex}
We use a custom matching algorithm based on regex patterns to extract the time and date from the model responses. For example, from a response like \emph{``In 5 hours it is 02:17 AM''}, we extract \emph{``02:17 AM''} and compare it to the gold standard answer. We chose to parse the model's response rather than enforce a strict output format. This allows us to keep prompts more natural and ensures that we do not penalize correct answers that only slightly deviate from the expected structure.

We develop and test the regex evaluator in two steps. First, we ask two independent annotators to evaluate a sample of 3,600 model responses for their correctness (\emph{true / false}) compared to the gold answers. Second, we split this data 50:50 into a development set, for constructing and optimizing regex patterns and a test set for evaluation. We achieve an almost perfect Cohen's $\kappa$ annotator agreement on the labeling task according to the scale of \citet{landis_koch}, with agreements in the range of 90.8-99.2\% across models, ensuring the quality of the obtained labels. Our regex evaluator achieves an accuracy of at least 98~\% for each model on the test set portion of the annotated subset. These scores confirm that we can use our regex evaluator for our experiments without introducing substantial noise into the evaluation results. Note also that the regex, with near-perfect accuracy, is vastly more efficient than training a response classifier or using LLM-as-a-judge. For further details on the validation process, see Appendix~\ref{app:regex_eval}.

%To evaluate the generated model responses, we developed a custom approach using regex matches. This was possible because the answer to be found is a data format whose structure is easy to extract (i.e. xx:xx AM|PM) and the model answers usually follow a very similar structure. Therefore, a very simple matching algorithm is applicable in this case and resource-intensive methods such as training a classifier or LLM as a judge could be avoided. To develop and validate our evaluation methodology, we first drew a stratified sample of the model responses from 100 responses per question type and evaluated model, resulting in a set of 3,600 labels. 

\section{How Well Do LLMs Reason About Time, Place, and their Combination?}\label{sec:reason}
For a reliable performance estimate on our dataset, we show aggregated results across all three instruction types: We use the majority vote per input instance, i.e., the answer obtained for at least two corresponding instructions. This procedure thus also reflects possible improvements through chain-of-thought prompting on more complex tasks. We present these combined results in Figure~\ref{fig:acc_overall} and show detailed results in Appendix \ref{app:model_accuracy}.

\subsection{Overall Results}
%The results show that while the \textsc{Verification} and \textsc{TimeTime} prompt types are feasible for most models, they show a general drop in performance when it comes to incorporating location-specific knowledge.

\begin{figure}
    \centering
    \includegraphics[width=\linewidth]{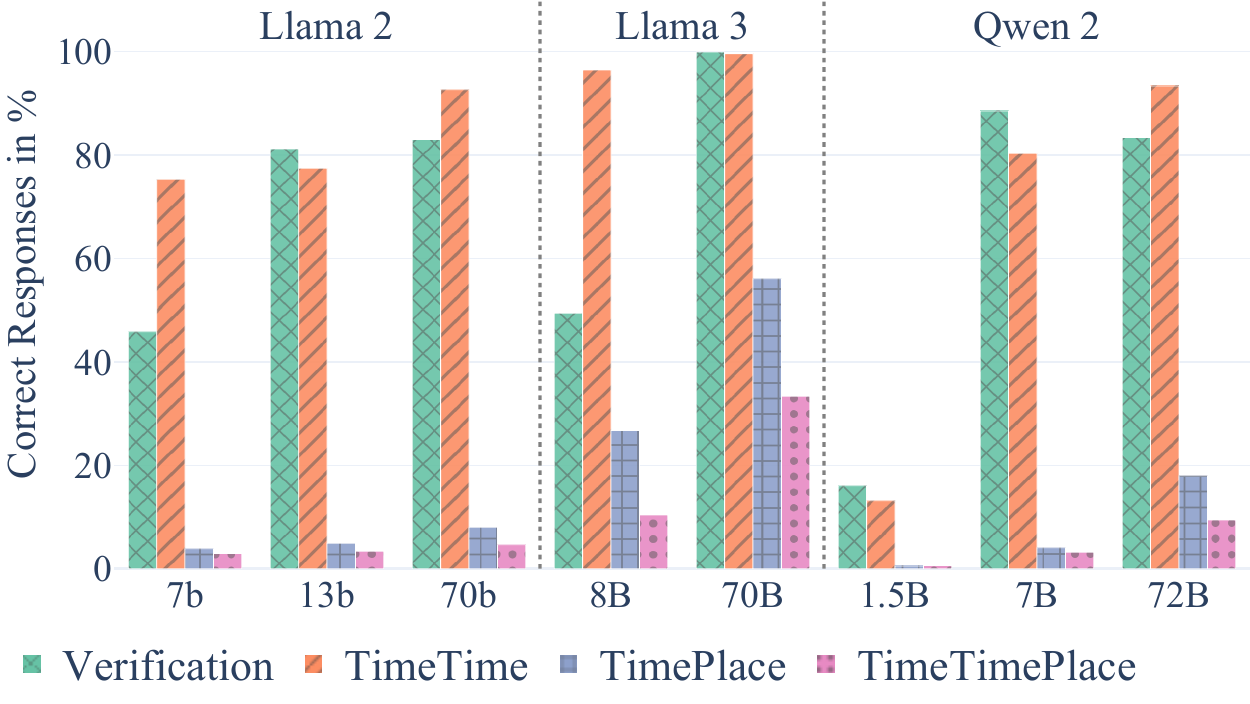}
    \caption{\textbf{Model accuracy on \textsc{GeoTemp}.} We show the percent of correct model responses (obtained via majority vote across different instruction types) for different models for our four geotemporal tasks (\textsc{Verification}, \textsc{TimeTime}, \textsc{TimePlace}, \textsc{TimeTimePlace}).}
    \label{fig:acc_overall}
\end{figure}

%The lowest score of 0.5\% is obtained by Qwen2-1.5B on \textsc{TimePlace} and the highest of 99.9\% by Llama3-70B on \textsc{Verification}. 
Performance varies drastically across models and tasks. Generally, while the \textsc{Verification} and \textsc{TimeTime} tasks are solvable for most models, they struggle when asked to incorporate location-specific knowledge. For instance,  Llama3-70B yields the best results with 56.1\% of correctly answered prompts for the \textsc{TimePlace} category and 33.4\% correct answers for the \textsc{TimeTimePlace} category. Interestingly, even though one would expect a monotonous performance decrease with higher task complexity, as for Llama2-13B and the smaller Qwen2 models, all other models seem to struggle with the \textsc{Verification} task while performing much better on the \textsc{TimeTime} task. We hypothesize this may be linked to ongoing training on mathematical problems, as LLMs have generally been shown to struggle with such tasks \cite{dziri2023faithfatelimitstransformers}. Very simple or even rhetorical questions, such as our \textsc{Verification} question, are probably less frequently included in the training data as they do not contain a common task to be solved. An analysis of several model responses shows that the models often attempt to read into the question a task that is not actually posed, such as providing further current times in other time zones.

\subsection{Model Scaling}
While we see consistently poor performance for all models on the two tasks that require geographical knowledge (\textsc{TimePlace} and \textsc{TimeTimePlace}), we also observe consistent scaling effects. For Llama2, this effect is more apparent for the two other and less complex tasks, while for Llama3 and Qwen2, a performance increase can be observed for all four tasks as the model size increases. In the 70B version, Llama3 thus achieves almost 100\% accuracy for the \textsc{Verification} and \textsc{TimeTime} task and a performance increase of 29.3\% for \textsc{TimePlace} and 23.0\% on \textsc{TimeTimePlace} compared to Llama3-8B. For Qwen2, however, an increase in model size results in only marginal improvements, with final performances of 18.0\% and 9.4\% on the complex tasks, respectively. For Llama2, the gains are even less apparent. This raises doubts as to whether the models will understand the underlying mechanisms for solving these tasks simply by upscaling.

\subsection{Instruction Types}
Previously, we analyzed aggregated results across the three instruction types (\emph{neutral}, \emph{chain-of-thought}, \emph{short}), but examining them individually also reveals interesting patterns. Forcing concise answers (\emph{short}) leads to decreased performance on complex tasks across all models, while improving results on simpler tasks, except for Qwen2.  In contrast, \emph{chain-of-thought} instructions result in significantly worse performance on simpler tasks for Llama2 and slightly worse for Llama3 compared to \emph{neutral}, particularly for Llama2-70B in the \textsc{Verification} task. Qualitative analysis of the model's answers suggests that the model is trying to solve a more difficult task than requested and ultimately gets stuck in its own explanations. These results are related to the work of \citet{sprague2024cotcotchainofthoughthelps}, who show that \emph{chain-of-thought} prompting primarily benefits math-related tasks. This highlights the need for a more comprehensive analysis of the effect of different instruction types on questions of varying difficulty, which we leave for future work. We provide detailed results in the Appendix \ref{app:model_accuracy}.

\subsection{Robustness}
\begin{figure}[h]

     \begin{subfigure}[b]{0.45\textwidth}
         \centering
         \includegraphics[width=\textwidth]{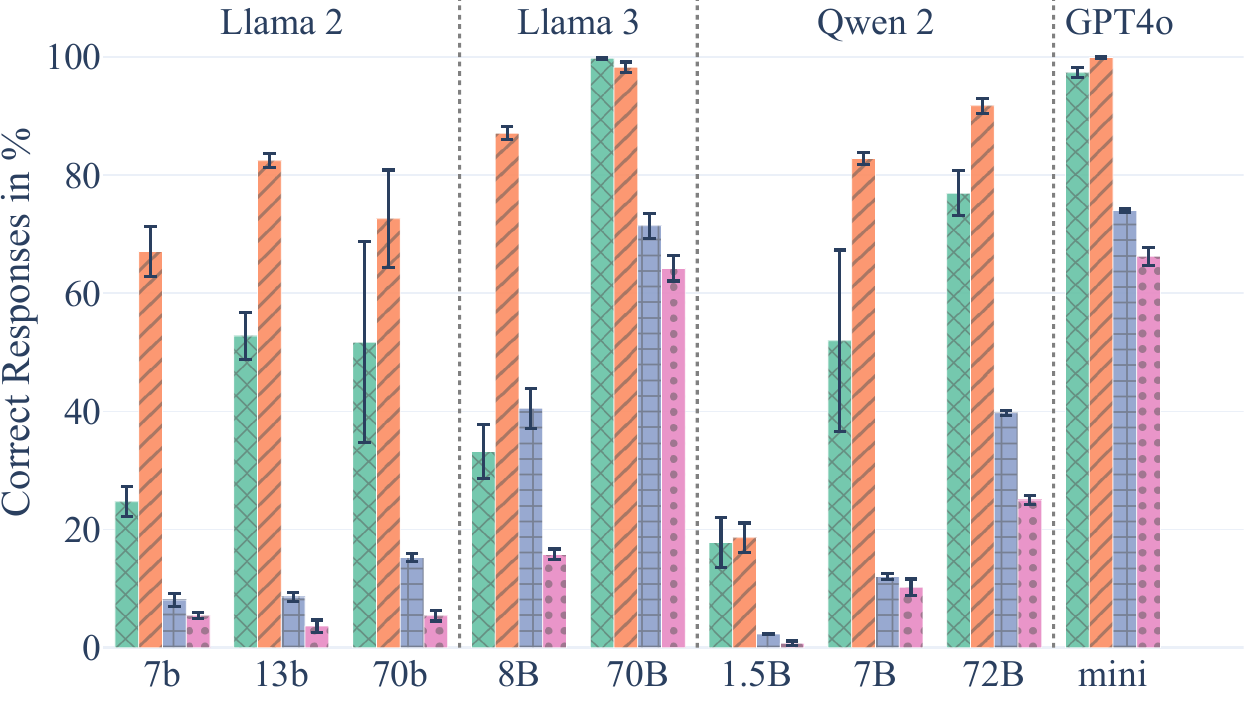}
         \caption{Robustness results for the \emph{neutral} instruction type.}
         \label{fig:robustness_a}
     \end{subfigure}

     \begin{subfigure}[b]{0.45\textwidth}
         \centering
         \includegraphics[width=\textwidth]{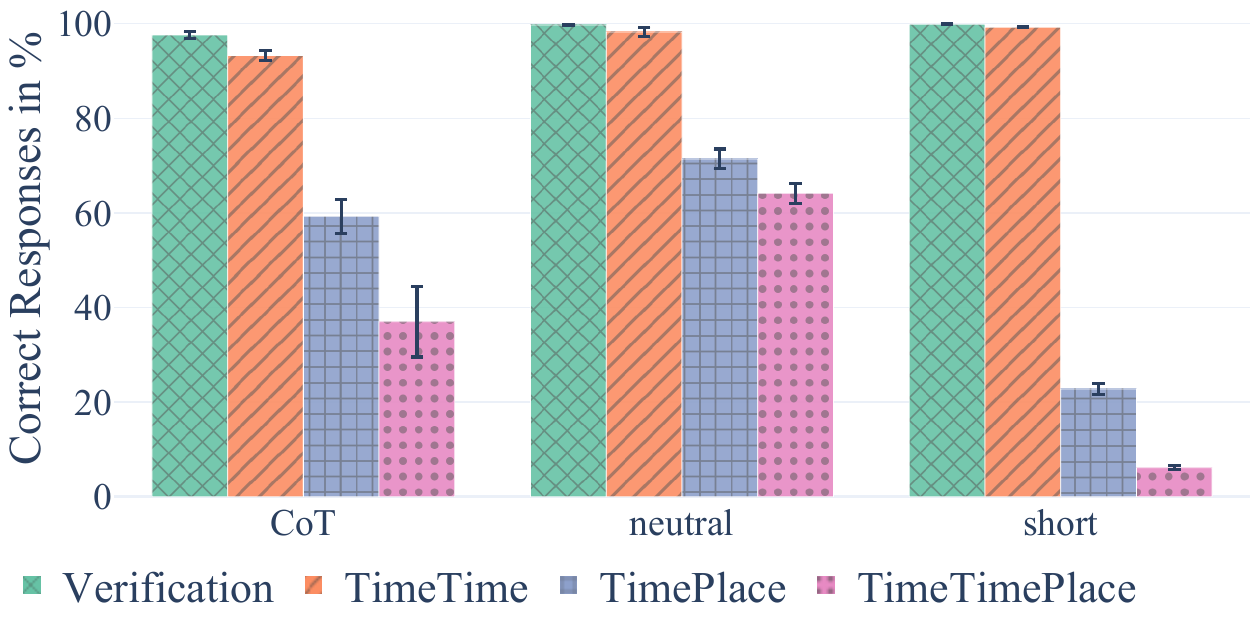}
         \caption{Llama3-70B results across instruction types.}
         \label{fig:robustness_b}
     \end{subfigure}
        \caption{\textbf{Robustness analysis.} We show the mean and standard deviation of model answers across three different prompt variations. In (a), the results for the \emph{neutral} instruction type across all models. In (b) we show the results obtained by Llama3-70B for each instruction type (\emph{neutral}, \emph{chain-of-thought}, \emph{short}).}
        \label{fig:robustness}
\end{figure}

Since the exact prompt formulation can have a strong impact on performance~\cite{sclarquantifying}, we additionally perform a robustness analysis. For each instruction type, we randomly sample 1,000 instances and generate results for two additional prompts with slight variations in wording. To enable comparison with state-of-the-art closed models, we also conduct the same experiment using \texttt{gpt4o-mini}. In Figure~\ref{fig:robustness_a}, we show the mean and standard deviation across all models for the instruction type \emph{neutral}, while Figure~\ref{fig:robustness_b} presents detailed results for all instruction types obtained with Llama3-70B. Results for the remaining models are provided in Appendix~\ref{app:model_accuracy}. While we observe significant performance fluctuations for some models and tasks, importantly, the overall trends between the models, their sizes, and instruction types are stable. Encouragingly, the error for the \textsc{TimePlace} and \textsc{TimeTimePlace} tasks, which we mainly analyze, is very low, highlighting the robustness of our findings.

%We show the mean and standard deviations obtained for Llama3-70B in Figure~\ref{fig:robustness_a} and provide the results for the other models in the Appendix \ref{app:model_accuracy}. {\color{black} For this experiment, we also report the results for \texttt{gpt4o-mini}, to provide a comparison to current SOTA closed models and find very similar patterns.} While we observe significant performance fluctuations for some models and tasks, importantly, the overall trends between the models, their sizes, and instruction types are stable. Encouragingly, the error for the \textsc{TimePlace} and \textsc{TimeTimePlace} tasks, which we mainly analyze, is very low, highlighting the robustness of our findings.

%\begin{figure}[t]
 %   \centering
%    \includegraphics[width=0.95\linewidth]{img/robustness_model_Meta-Llama-3-70B-Instruct.pdf}
%    \caption{\textbf{Robustness analysis.} For each  instruction type (\emph{neutral}, \emph{chain-of-thought}, \emph{short}), we show the mean and standard deviation across three different prompt variations for answers obtained by Llama3-70B.}
%    \label{fig:robustness}
%\end{figure}

\section{What Factors Influence Model Performance?}
\label{sec:factors}
\subsection{Country Biases}
\begin{figure}[t]
    \centering
    \includegraphics[width=\linewidth]{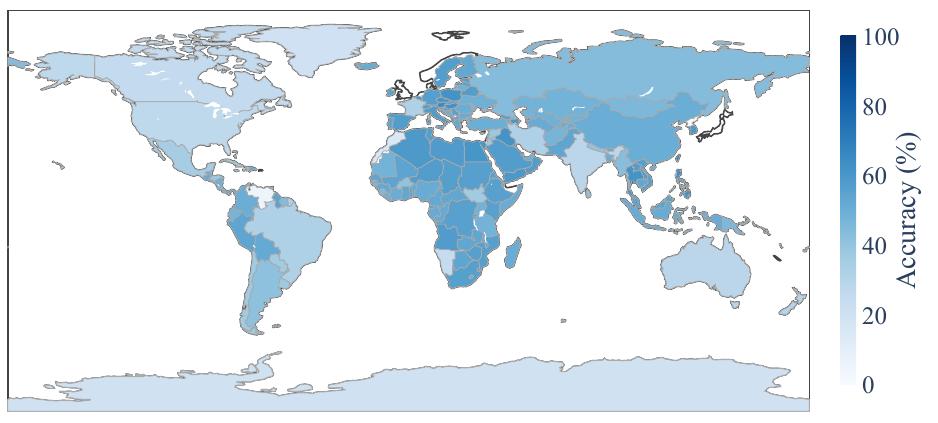}
    \caption{\textbf{Model accuracy aggregated by target country} for the tasks \textsc{TimePlace} and \textsc{TimeTimePlace}. We neither observe biases towards Western nor towards higher-income or population-dense countries.}
    \label{fig:model_acc_per_country}
\end{figure}

Following \citet{worldbench}, we investigate performance differences based on location, comparing Western vs.\ Non-Western countries, varying income levels, and population densities. For this, we focus on the tasks requiring location-specific knowledge (\textsc{TimePlace} and \textsc{TimeTimePlace}), aggregating results by start and target country. Figure \ref{fig:model_acc_per_country} depicts the response accuracy aggregated by the target country of Llama3-70B. We provide results for the aggregation by start country in Appendix \ref{app:country_bias}, which show no significant differences. 

Llama3-70B performs best on questions that include African countries. Similarly, the model performs well for cities in South America and around Russia. For cities in North America, Oceania, Antarctica, and Greenland, on the other hand, we see significantly worse results. This is contrary to our hypothesis that models develop a performance bias towards Western countries. Further aggregations at the continental level, population size, and income level did not reveal any clear patterns in this direction either.

\subsection{Location Name Perplexity}
Since we could not identify a clear location bias in the models, we explored other factors that might influence their performance. 

\paragraph{Detailed setup} We hypothesize that the familiarity of the models with certain locations leads to them performing better in a task. Since we cannot access the models' training data, we make use of the models' perplexity as a proxy to assess their familiarity with specific locations. %As measuring the perplexity for the name of the location alone is prone to error, 
To this end,  we use a simple and static sentence template: \textit{``I live in \{city\}, \{country\}}'', replace the placeholders with the corresponding city and country information for each location $l$ from \textsc{GeoTemp} and compute the perplexity for each model. To better interpret the perplexity levels, we categorize the resulting perplexity scores into four bins. Specifically, for each model, we determine the perplexity distribution over all locations in our dataset and classify locations with a perplexity score within the 25\% quantile of this distribution as \emph{``very low''} perplexity, those up to the 50\% quantile as \emph{``low''}, up to the 75\% quantile as \emph{``high''} and the remaining values as \emph{``very high''}. We then associate these with the corresponding start and target locations in our dataset and compute the average model performance in percent of correct answers for each combination of perplexity scores of the two locations in question. Note that, again, we are limiting this analysis to the \textsc{TimePlace} and \textsc{TimeTimePlace} tasks, where location information is relevant.

\begin{figure}[t]
    \centering
    \includegraphics[width=0.7\linewidth]{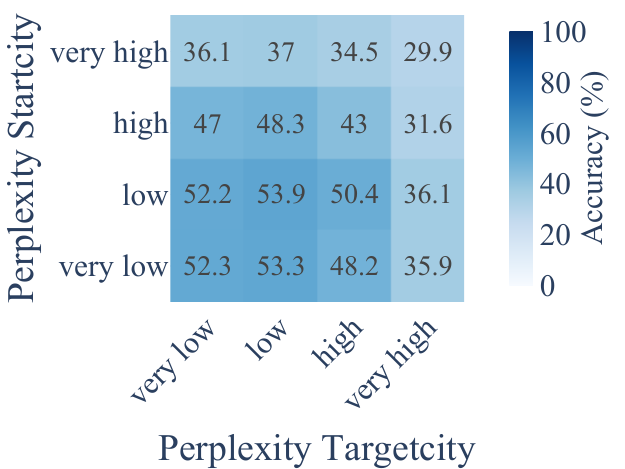}
    \caption{\textbf{Effect of location perplexities.} We show the performance of Llama3-70B on the \textsc{TimePlace} and \textsc{TimeTimePlace} task aggregated by the perplexity of the start location and target location.}
    \label{fig:ppl_heat_llama3}
\end{figure}

\paragraph{Results} We focus on the result of  Llama3-70B, shown in Figure \ref{fig:ppl_heat_llama3}, because of its strong performance across tasks. However, we find that these findings also hold for most other models which we show in the Appendix \ref{app:perplexity}. %We find that all models except Qwen2-1.5B achieve the best model performance for combinations of locations for which the model exhibits \emph{very low} or \emph{low} perplexity. Note that Qwen2-1.5 exhibits very poor overall performance for the two tasks in question, generating correct answers rather randomly. 
For Llama3 we observe a gradual decrease in accuracy from \emph{low} to \emph{very high} perplexity locations, with a surprisingly high performance drop between the combination of \emph{low x low} (53.9) and \emph{very high x very high} (29.9\%) of 22.5\%. This trend is also visible for Qwen2-72B with a performance difference of 15.5\% between the combinations \emph{very low x very low} and \emph{very high x very high}. Our findings thus suggest that model performance is biased in favor of locations that likely appear more frequently in model training data, rather than being biased \emph{per se} in the direction of Western countries, as one might have expected.

%Interestingly, there is indeed a connection between the model perplexity of the locations and its performance for the time-shift tasks. 

\subsection{Error Analysis}
\begin{figure}[t]
    \centering
    \includegraphics[width=\linewidth]{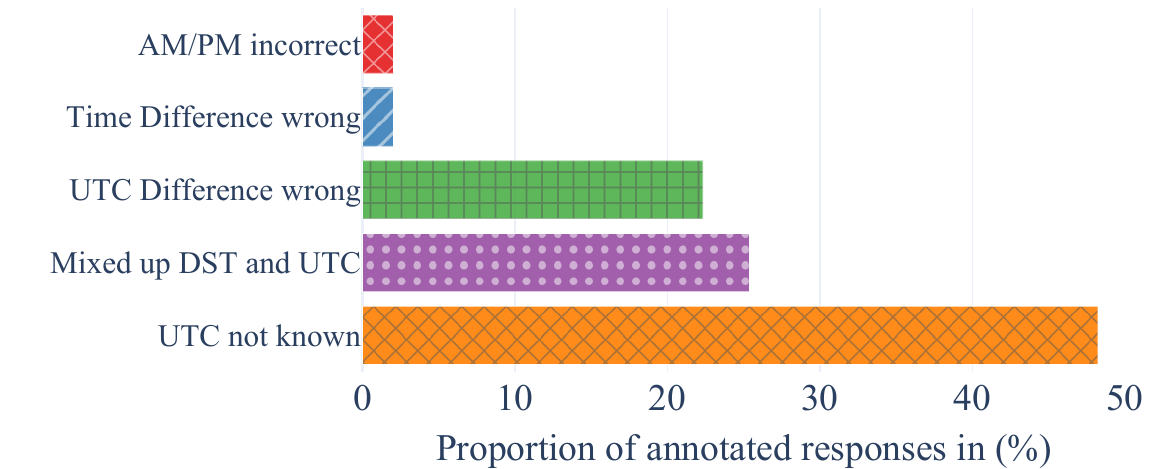}
    \caption{\textbf{Results of our error analysis.} We show the proportion of model errors (in $\%$) per error category.}
    \label{fig:qualit_analysis}
\end{figure}

To get more precise insights into why models fail to combine temporal and geographical knowledge, we perform a manual error analysis for the best-performing model, Llama3-70B. We randomly sample 200 of the model's incorrect responses on the \textsc{TimeTimePlace} task.
One author then annotated this sample, assigning errors to one of five error categories (Figure~\ref{fig:qualit_analysis}). We find that 25.3\% of errors are attributable to the model's failure to appropriately handle the conversion between DST and UTC for the current time. 22.3\% of errors result from the fact that, despite correct UTC knowledge for both locations, the time difference was not calculated correctly. However, 48.2\% of the errors are due to the fact that the model does not use the correct UTC in its decision-making process for at least one of the locations.

\subsection{Location-Specific Knowledge}
We seek to further investigate why models perform so poorly on tasks requiring a combination of temporal and geographical knowledge. From our \textsc{TimeTime} task, it is evident that models are generally able to perform basic time calculations, hence we look at the time zone knowledge of the models for the locations. To do this, we prompt the models to provide us with the corresponding time zone for each location in our dataset. For an easier parsing of the results, we additionally force the model to respond following the scheme: \emph{``City: UTC±X''}. The outcomes of this evaluation, presented in Table \ref{tab:utc_match}, are surprising in light of the previous results: All models can predict the correct UTC timezone for at least 65\% of the locations, except for Qwen2-1.5B with only 39.3\%. For Llama3-70B, it is as much as 90\%. Consequently, the poor results from our main experiments (Figure \ref{fig:acc_overall}) cannot be attributed solely to a lack of knowledge about the cities and their geographical locations. Rather, models fail to make use of this knowledge when not explicitly asked to provide it.

\setlength{\tabcolsep}{18pt}
\begin{table}[t]
    \centering
    \small
    \begin{tabular}{lrr}
    \toprule
    \textbf{Model} & \textbf{Size} & \textbf{UTC Known ($\%$)} \\
    \midrule
            &   7B &   70.9  \\
      Llama2 &  13B &    71.3 \\
      &  70B &    80.9 \\
\midrule
     \multirow{2}{*}{Llama3} &   8B &   84.1\\
            & 70B&     \textbf{90.0}   \\
  \midrule
       &  1.5B &    39.3 \\
       Qwen2 &  7B &   65.4  \\
       &  72B &    86.3 \\
       
\bottomrule
\end{tabular}
    \caption{\textbf{Knowledge about time zones}. We present the proportion of locations where models assigned the correct time zone (in $\%$ of all locations in our dataset) when explicitly prompted to do so.}
    \label{tab:utc_match}
\end{table}

%After enriching our data set with the information obtained and adding the information whether the model knew the UTC time zone of the respective start and target location involved in the prompt in our previous test.

\section{Can Model Performance be Improved by Injecting Geotemporal Knowledge?} \label{sec:knowledge_inj}
\paragraph{Detailed setup}
So far, we found that the models are generally able to perform temporal calculations and that knowledge about the individual locations is present, yet they fail when combining the two. In a final experiment, we now examine whether explicitly injecting geographical knowledge by providing time zone information helps to activate the required knowledge (and thus, improve performance on the most complex task \textsc{TimeTimePlace}). %In doing so, we reduce the complexity of the task by relieving the model of the first step, the determination of the UTC timezones, and also hint at the correct processing procedure. 
We examine two different setups: In the first setup, we add the respective time zone information of the city in addition to the input prompt (\emph{``add time zone''}). In the second variant, we omit the city names from the prompt and replace them entirely with the time zone information (\emph{``replace by time zone''}). 

\begin{figure}[t]
    \centering
    \includegraphics[width=\linewidth]{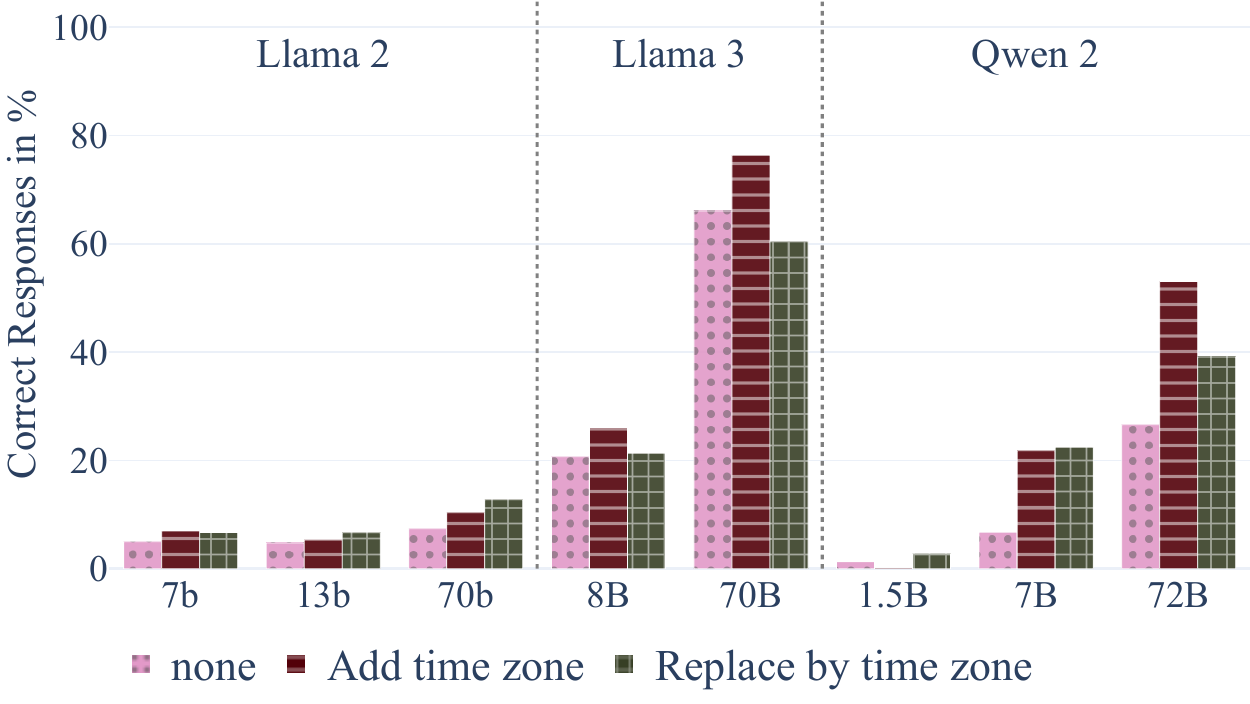}
    \caption{\textbf{Results of UTC knowledge injection} on the \textsc{TimeTimePlace} task. Colors respond to different amounts of knowledge injected.}
    \label{fig:utc_injection}
\end{figure}

\paragraph{Results}
We present model performance with injected information in comparison to prior performance on the \textsc{TimeTimePlace} task in Figure~\ref{fig:utc_injection}. We see that the injection of geotemporal knowledge improves performance across models. For Llama2, however, improvements are limited, pointing to a general lack of understanding of the concept of time zones. For the large versions of the other models, we see a significantly larger effect. Llama3-70B achieves a performance of 76.3\% by adding the time zone information, and Qwen2-72B of 53\%.
We also find that models do not perform as well when city information is replaced with time zone information as when time zone information is added on top. This suggests that models do draw on their geographical knowledge to solve this task.

%\paragraph{Alignment of Results}

%\begin{table}[]
%    \centering
%    \small
%    \begin{tabular}{l|ccc}
%    \toprule
%      \multirow{2}{*}{\textbf{Strategy}}   & \multicolumn{3}{c}{\textbf{Number of UTCs known before}}  \\
%         & \multicolumn{1}{c}{none} & \multicolumn{1}{c}{one} & \multicolumn{1}{c}{both} \\
%         \midrule
%       Only UTC & \multicolumn{1}{c|}{1.8} & \multicolumn{1}{c|}{25.6} & \multicolumn{1}{c}{72.7} \\
%        UTC and City & \multicolumn{1}{c|}{1.5} & \multicolumn{1}{c|}{24.4} & \multicolumn{1}{c}{74.1} \\
%         \bottomrule
%    \end{tabular}
%    \caption{Proportion of new correct responses by adding UTC and how many UTCs the models have known before injecting that knowledge.}
%    \label{tab:alignment_utc}
%\end{table}

\section{Discussion}
%Ultimately, we combine our findings to highlight key insights. 
{\color{black} Current language models struggle to perform geotemporal reasoning. Our dataset and analysis allowed us to identify reasons for this.} First, we find that the only model with fewer than 7 billion parameters in our test (Qwen2-1.5B) is largely ineffective for geotemporal reasoning as required by the tasks in our dataset. In contrast, we show that bigger models can indeed perform basic temporal calculations, indicating a solid understanding of temporal concepts like the 24-hour format and AM/PM notation. Especially, the example of Llama3-70B, with almost 100\% accuracy on this task (\textsc{TempTemp}), shows that targeted training (e.g., high-quality data and specific training for STEM tasks) can achieve a significant performance improvement. 
Second, most models – except for Qwen2-1.5B – correctly determine time zones for at least 65\% of the locations when explicitly asked to do so, demonstrating general geographic competencies. Third, despite this, their performance on tasks combining temporal and geographical knowledge is poor. Our qualitative analysis reveals that when models have to perform complex tasks that require them to access and combine different types of knowledge simultaneously, they are no longer able to retrieve time zone information that they knew in isolation and instead start hallucinating. Our knowledge injection tests support this finding, as reducing task complexity by limiting the amount of information to be retrieved at once significantly improves model performance. This analysis points to a broader issue: \textbf{models may often possess the necessary knowledge but fail to retrieve and combine it effectively for complex tasks} (and steering techniques like \emph{chain-of-thought} may only partially help). Looking into the future, we also believe that simply increasing the model size will not solve the problem. Instead, improvements towards more systematic and step-wise knowledge retrieval or the use of tools may be necessary to prevent hallucinations and sustainably enhance performance on complex reasoning tasks.

\section{Conclusion}
In this article, we introduced \textsc{GeoTemp}, the first benchmark designed to evaluate the ability of LLMs to reason over both temporal and geographical knowledge at different combinations of knowledge. \textsc{GeoTemp} consists of over 320k prompts covering 289 locations across 37 timezones. Using our dataset, we analyzed eight open chat-optimized LLMs across varying sizes and found that, while models perform adequately on tasks involving only temporal knowledge, their performance significantly drops when they are required to combine temporal and geographical information. Our findings indicate that this limitation persists also at larger model scales, suggesting that even the best models today struggle with more complex reasoning tasks involving the interplay of time and space. We also observed that the models perform better for locations where they exhibit low perplexity, which may indicate that prior exposure to certain geographic locations during training helps.

\section*{Limitations}
\paragraph{Regex Evaluation}
We use a self-developed algorithm to evaluate the open-ended model responses, which inevitably causes some degree of noise. This was necessary as a full human evaluation for a dataset of this size would not be feasible, and LLM as a judge would be too resource-intensive and also not fully reliable. However, we are conducting a comprehensive validation of our methodology, providing very high accuracy, which leads us to believe that we can consider the noise to be low. We are therefore confident that our results remain valid. 

\paragraph{Prompt Robustness Evaluation on Selected Subset}
Due to resource constraints, we conduct robustness checks only on a subset of our dataset. It is therefore possible that the true uncertainties around our results are larger. Third, in our experiments, we evaluate models for which we unfortunately do not have any precise information about their pretraining data. To make a statement about the model's familiarity with certain terms, we thus use model perplexity as a proxy for the frequency of the term's occurrence in the training data. Yet, this metric is not fully reliable and may introduce inaccuracies. A future evaluation of fully open models and their training data could provide additional insights. 

\paragraph{No Tasks Involving Location Alone}
\textsc{GeoTemp} does not cover purely location-based tasks, as such a task would closely resemble existing work and might not significantly advance the field. Furthermore, a task that involves predicting a location is less conducive to leveraging temporal information compared to the task we have chosen to focus on.

\section*{Acknowledgements}
Paul Röttger is a member of the Data and Marketing Insights research unit of the Bocconi Institute for Data Science and Analysis, and is supported by a MUR FARE 2020 initiative under grant agreement Prot. R20YSMBZ8S (INDOMITA). The work of Carolin Holtermann and Anne Lauscher is funded by the Excellence Strategy of the German Federal Government and the Federal States.

\clearpage

% Bibliography entries for the entire Anthology, followed by custom entries
%\bibliography{anthology,custom}
% Custom bibliography entries only
\bibliography{custom}

\clearpage

\appendix

\addcontentsline{toc}{section}{Appendix} % Add the appendix text to the document TOC
\part{Appendix} % Start the appendix part
\parttoc % Insert the appendix TOC

\onecolumn
\section{Dataset} \label{app:dataset_dist}
\subsection{Sources}

We present a list of datasets and frameworks, we use for creating GeoTemp. The timezone information is obtained from the OTZD which we access using the Python libraries \texttt{pytz} and \texttt{datetime}.

\begin{table*}[h]
    \centering
    \small
    \begin{tabular}{cp{3cm}p{6cm}c}
    \toprule
    Purpose & Name  &  Source & License\\
    \midrule
      &  Country population,    & Worldbank \url{https://data.worldbank.org/} & CC BY 4.0\\
      Datasets&  Country names,     & & \\
      & Income Level, Regions    & & \\
       & & & \\
     &City population & Geonames/Opendatasoft \url{https://public.opendatasoft.com/explore/dataset/geonames-all-cities-with-a-population-500/api/?disjunctive.country} & CC BY 4.0\\
     \midrule
     Frameworks & pytz==2024.1 & \url{https://pypi.org/project/pytz/} &\\
      & python-dateutil==2.8.2 & \url{https://docs.python.org/3/library/datetime.html} &\\
    \bottomrule
    \end{tabular}
    \caption{Overview of datasets and frameworks used in our work.}
    \label{tab:licenses}
\end{table*}

\clearpage

% Time zone data is accessible through Python using the pytz package\footnote{\url{https://pypi.org/project/pytz/}}, in combination with the datetime package\footnote{\url{https://docs.python.org/3/library/datetime.html}}, facilitating a seamless integration for efficient transformation and evaluation of time data.

\twocolumn
\subsection{Quantitative Analysis}

\begin{figure}[h]
    \centering
    \includegraphics[width=\linewidth]{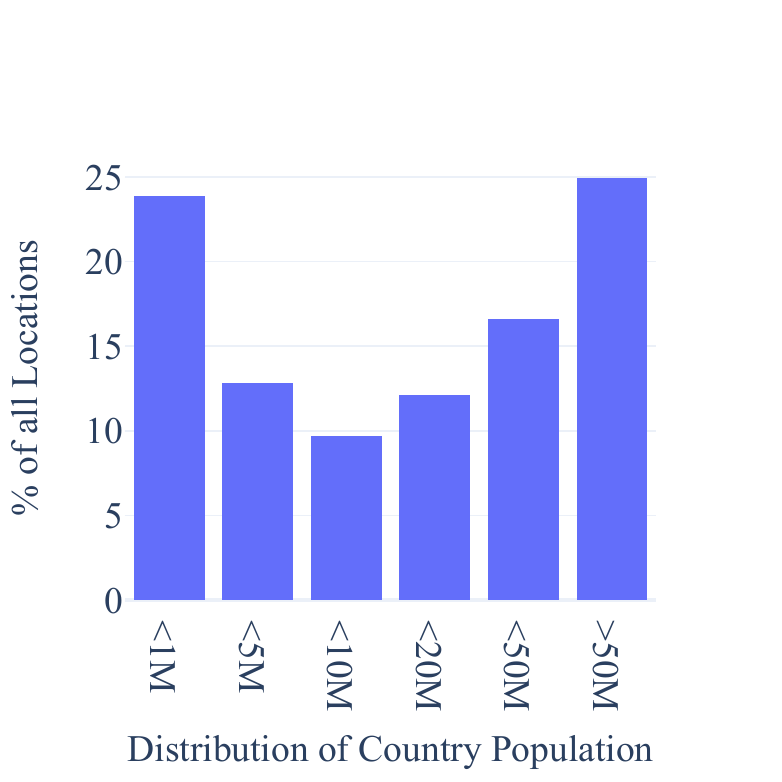}
    \caption{Histogram Country Population}
\end{figure}

\begin{figure}[h]
    \centering
    \includegraphics[width=\linewidth]{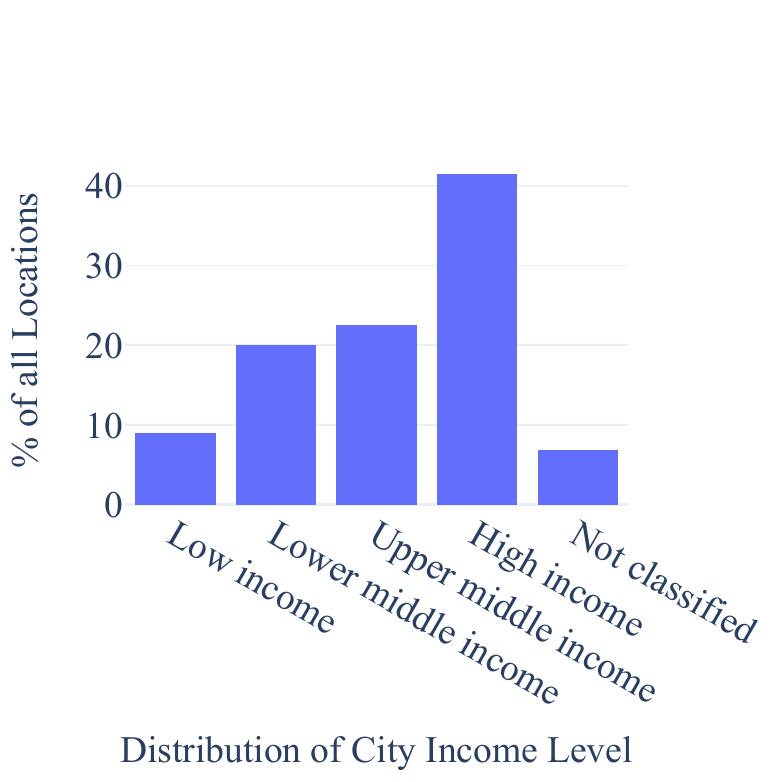}
    \caption{Histogram Country Income Level}
\end{figure}

\begin{figure}[h]
    \centering
    \includegraphics[width=\linewidth]{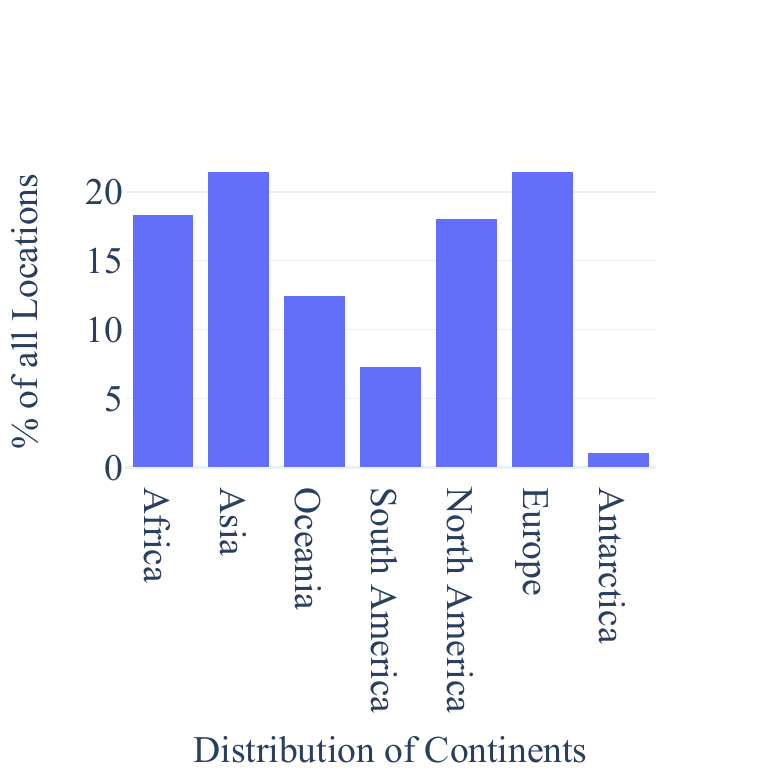}
    \caption{Histogram Continents of Locations}
\end{figure}

\begin{figure}[h]
    \centering
    \includegraphics[width=\linewidth]{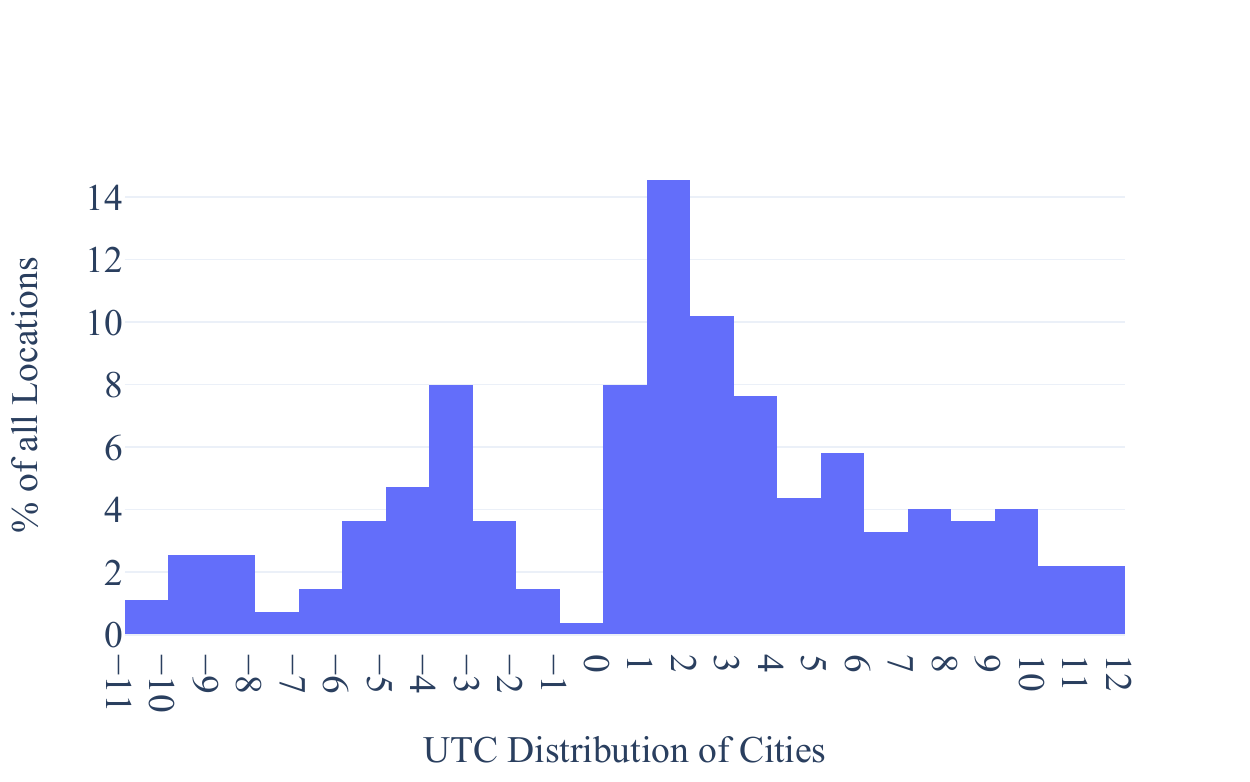}
    \caption{Histogram UTC of Locations}
\end{figure}

\clearpage

\section{Evaluation Setup} 
\label{app:regex_eval}

\paragraph{Annotation} 
We have two independent human annotators label this set of answers for their correctness with regard to the target answer given by the \textit{pytz} package. Both annotators are fluent in English and are familiar with the task at hand. Since the annotators both worked in the authors' research environment, they did not receive any compensation. They were also informed about how their annotations would be used. 

Expectedly, we achieve an overall agreement of 96.4\% of Cohen's $\kappa$ on the annotation set, constituting a very high agreement between the annotators and thus very reliable labels. In a qualitative analysis, it can be seen that most disagreement was caused by either random errors or on the \texttt{\textsc{Verification}} task. This is because models are often confused by this rather rhetorical type of question, and instead of returning the answer, they convert the time at a location into the corresponding time in UTC. This is generally not a wrong answer, but we specify for the annotation guidelines that the model must also respond with the correct time at the respective location. To obtain the final label, disagreements between the annotators were overruled by an expert decision judgment.

\begin{table}[h]
    \centering
    \small
    \begin{tabular}{llr}
    \toprule
    \textbf{Model} & \textbf{Size} & \textbf{Cohen's $\kappa$} \\
    \midrule
            &   7B &   90.8  \\
      Llama2 &  13B &    94.3 \\
      &  70B &    97.5 \\
\midrule
     \multirow{2}{*}{Llama3} &   8B &   97.5\\
            & 70B&     99.2   \\
  \midrule
       &  1.5B &    96.9 \\
       Qwen2 &  7B &   97.4  \\
       &  72B &    97.4 \\
       
\bottomrule
\end{tabular}
    \caption{\textbf{Inter-annotator agreement} on a sample of 3,600 model answers from GeoTemp. We present the Cohen's $\kappa$ between the two annotators in percent.}
    \label{tab:iaa_agreement}
\end{table}

\paragraph{Validation}
Next, we split the annotated subset using a 50:50 split into a calibration and a test portion. Note that, since we are not actually training a new model but rather using the set to calibrate our own matching algorithm, a larger test portion to validate the generalization of our approach on unseen data is more important than the calibration portion. The evaluation results for our matching algorithm compared to the human annotations are shown in Table \ref{tab:testset_performance}.

\setlength{\tabcolsep}{10pt}
\begin{table}[h]
    \centering
    \small
    \begin{tabular}{llrrr}
    \toprule
    \textbf{Model} & \textbf{Size} & \textbf{Acc} &  \textbf{P} &  \textbf{R} \\
    \midrule
            &   7B &   99.5 &       98.4 &   100.0 \\
      Llama2 &  13B &    98.0 &       95.9 &    98.6 \\
      &  70B &    98.0 &       98.9 &    96.7 \\
\midrule
     \multirow{2}{*}{Llama3} &   8B &   98.5 &       97.8 &    98.9 \\
            & 70B&     98.5 &      100.0 &    98.2 \\
  \midrule
       &  1.5B &    99.5 &       94.7 &   100.0 \\
       Qwen2 &  7B &    98.0 &       97.5 &    97.5 \\
       &  72B &    99.0 &      100.0 &    98.3 \\
\bottomrule
\end{tabular}
    \caption{\textbf{Performance of our regex evaluator.} Evaluation results on the sample test set of our dataset. We present the accuracy, precision, and recall for each model in percent on an annotated subset of 100 prompts each.}
    \label{tab:testset_performance}
\end{table}

The results show, that we can achieve an accuracy of at least 98\% on the test set portions for each model using our algorithm. The lowest precision score of 94.7\% was achieved for the 1.5B version of Qwen2. This is due to the fact that the model provides very few correct answers, and thus the true positive rate is very low, causing the precision score to decrease rapidly even though there is only one misclassified instance. Our validation results therefore suggest that our evaluation method delivers reliable results for the succeeding analysis.

\clearpage
\onecolumn
\section{Experimental Results} \label{app:model_accuracy}
\subsection{Robustness Analysis} 

We present the accuracy results on the whole dataset and the robustness results on a subset of 1000 prompts with three different prompt variations across the three different instruction types: (1) neutral, (2) short, and (3) CoT. We prompt \textsc{gpt-4o-mini-2024-07-18} in the same way as the other models, which we detailed in Section \ref{sec:experiments} via the OpenAI API\footnote{\url{https://platform.openai.com/docs/overview}}.

\begin{figure}[h]
     \centering
     \begin{subfigure}[b]{0.45\textwidth}
         \centering
         \includegraphics[width=\textwidth]{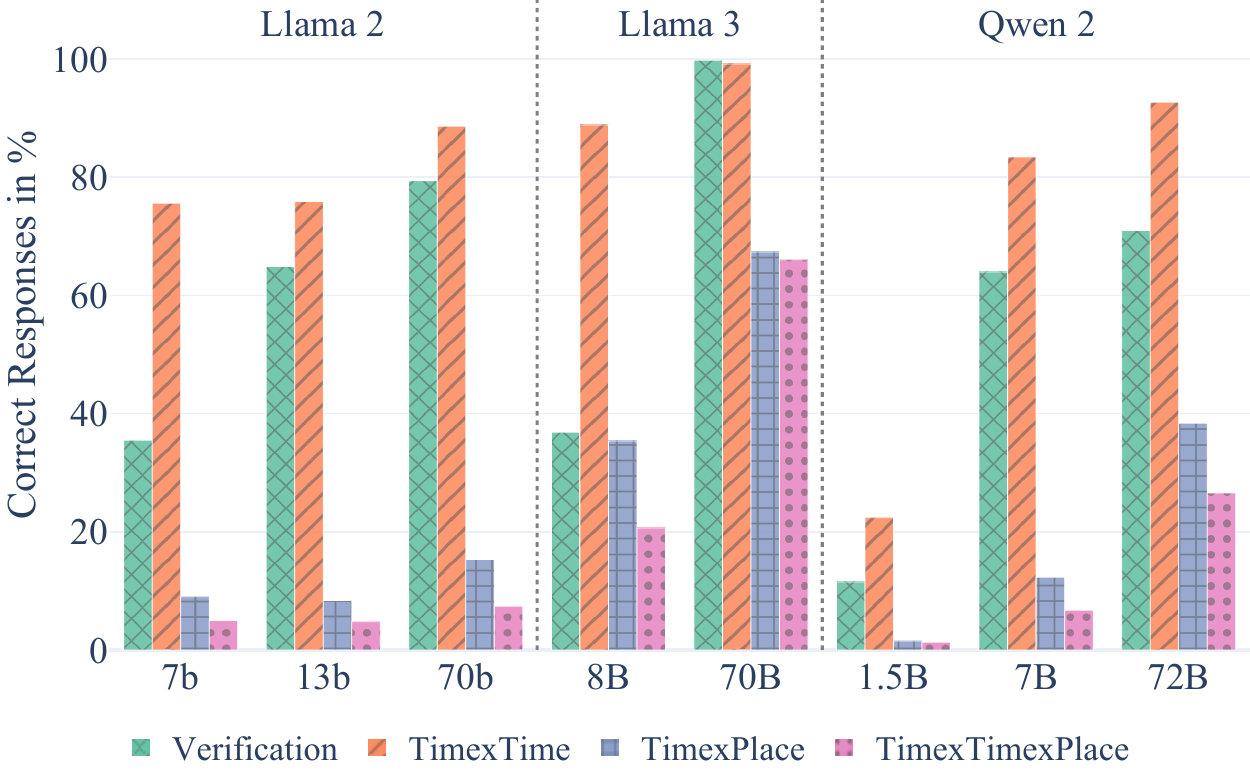}
         \caption{Overall Results}
     \end{subfigure}
     \begin{subfigure}[b]{0.45\textwidth}
         \centering
         \includegraphics[width=\textwidth]{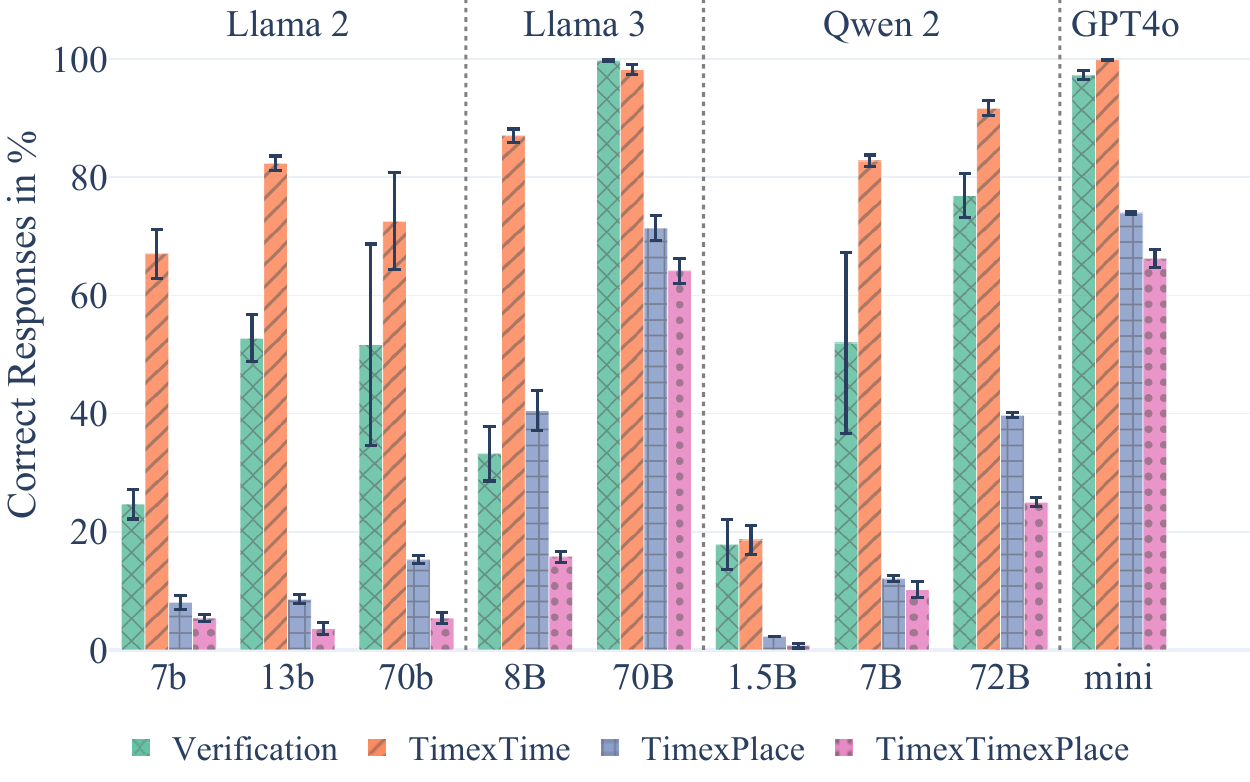}
         \caption{Robustnesschecks}
     \end{subfigure}
        \caption{Model Accuracy neutral prompt}
        \label{fig:acc_prompt1}
\end{figure}

\begin{figure}[h]
     \centering
     \begin{subfigure}[b]{0.45\textwidth}
         \centering
         \includegraphics[width=\textwidth]{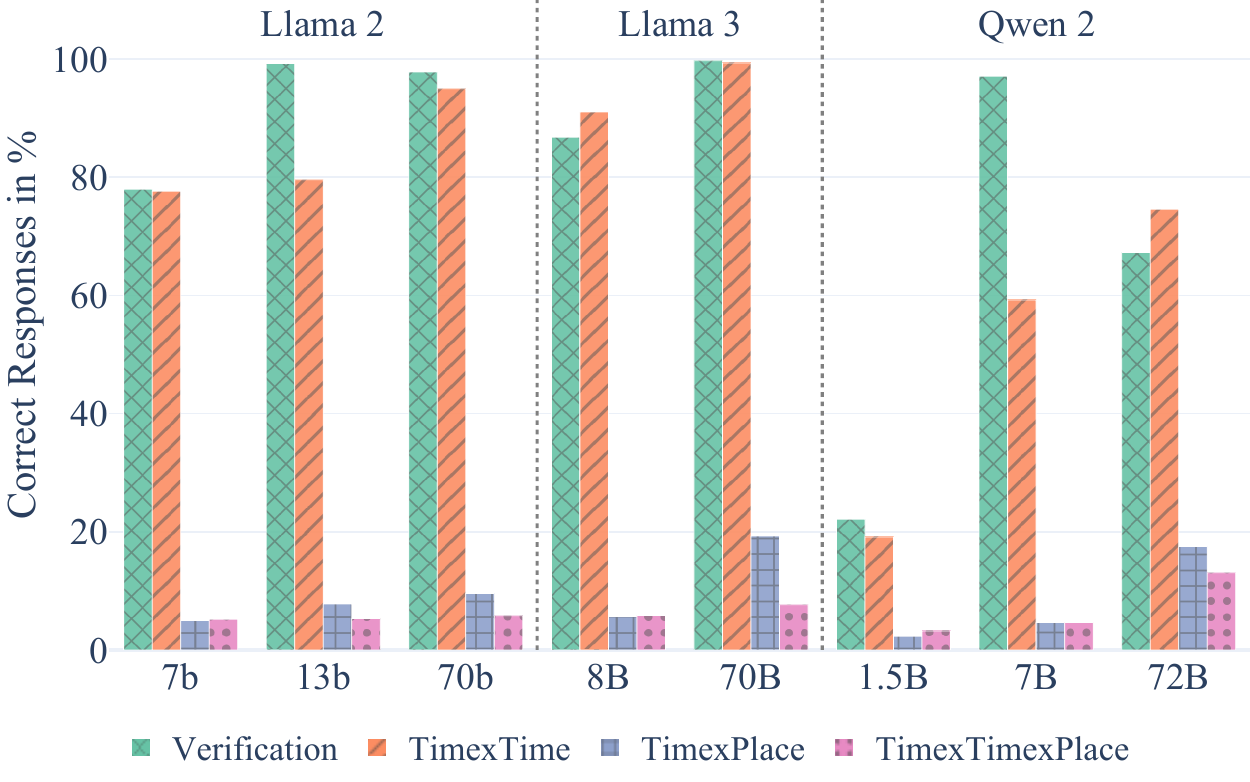}
         \caption{Overall Results}
     \end{subfigure}
     \begin{subfigure}[b]{0.45\textwidth}
         \centering
         \includegraphics[width=\textwidth]{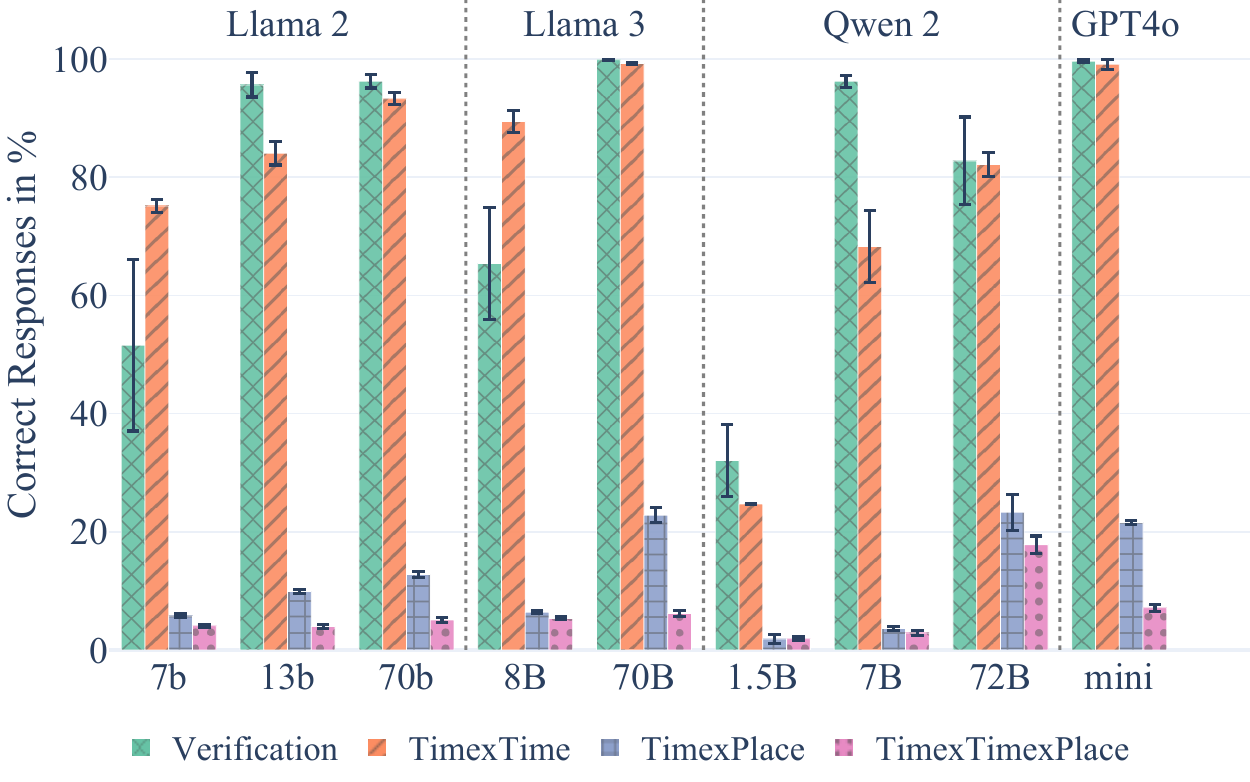}
         \caption{Robustnesschecks}
     \end{subfigure}
        \caption{Model Accuracy short prompt}
        \label{fig:acc_prompt2}
\end{figure}

\begin{figure}[h]
     \centering
     \begin{subfigure}[b]{0.45\textwidth}
         \centering
         \includegraphics[width=\textwidth]{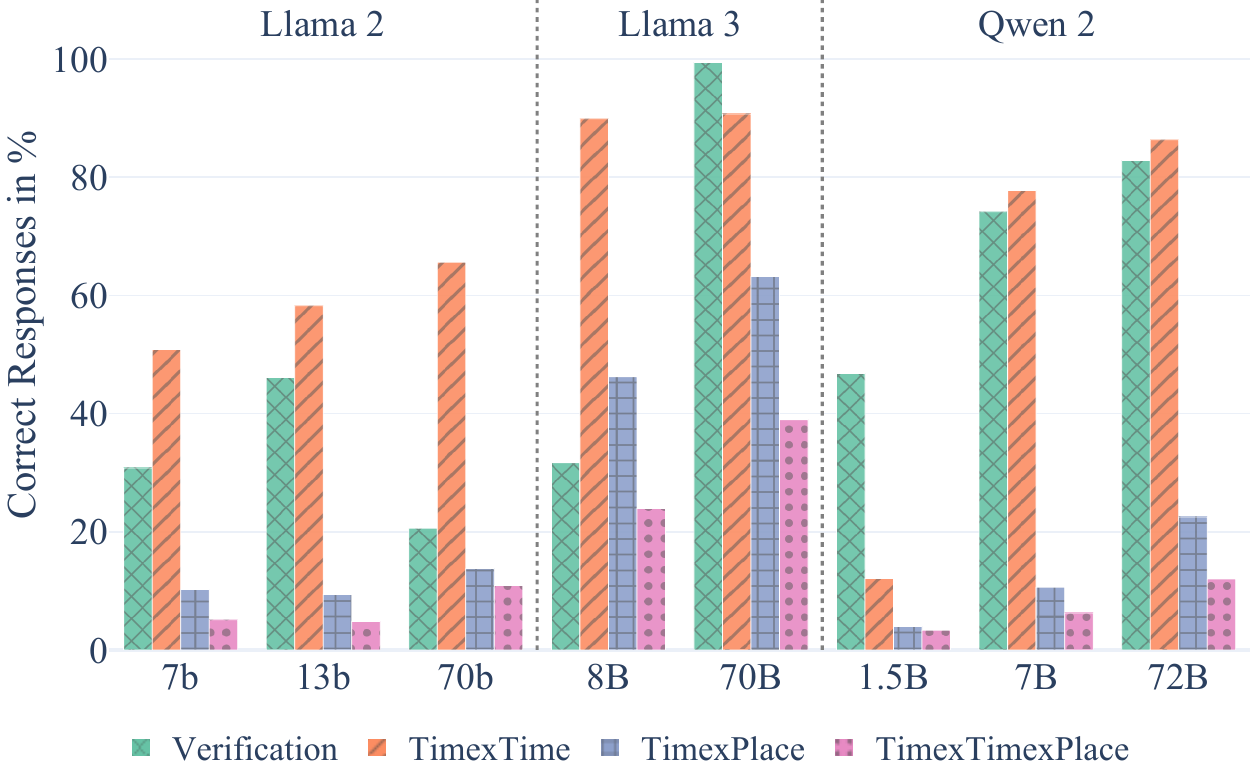}
         \caption{Overall Results}
     \end{subfigure}
     \begin{subfigure}[b]{0.45\textwidth}
         \centering
         \includegraphics[width=\textwidth]{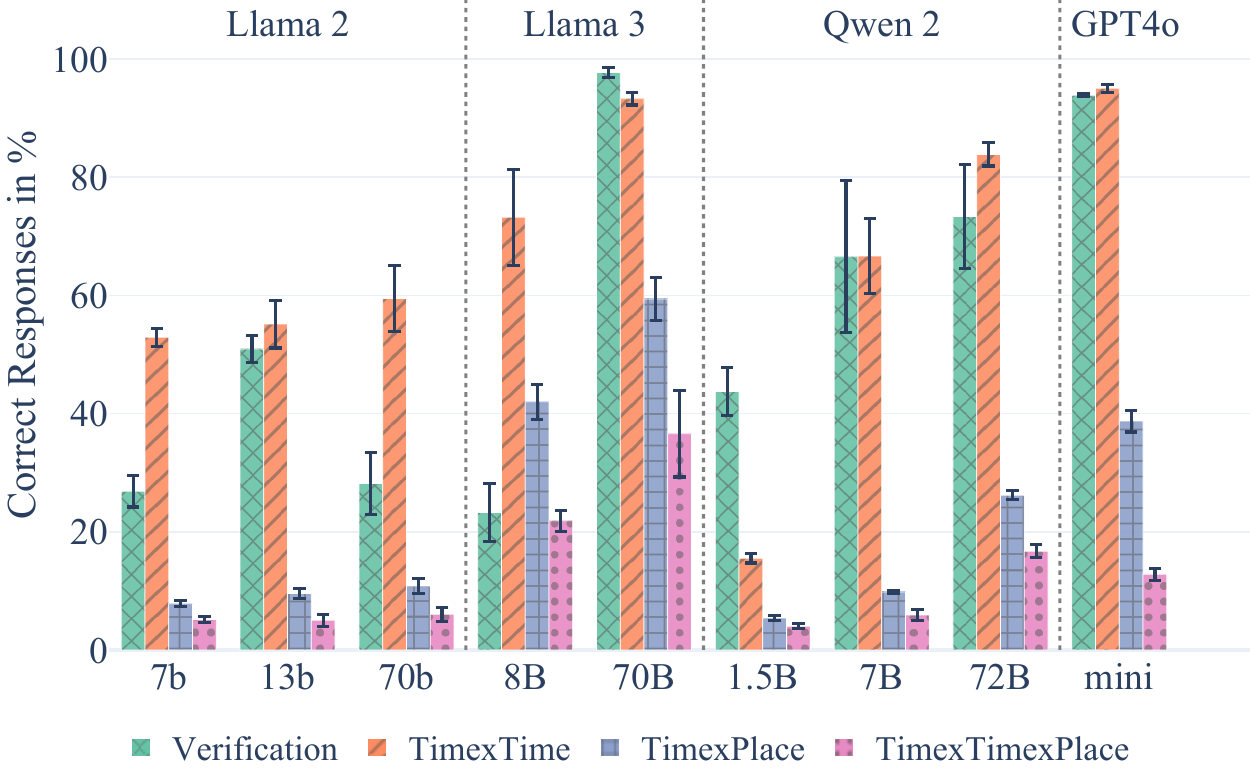}
         \caption{Robustnesschecks}
     \end{subfigure}
        \caption{Model Accuracy CoT prompt}
        \label{fig:acc_prompt3}
\end{figure}

\clearpage
\subsection{Country Bias Analysis} \label{app:country_bias}

\begin{figure}[h]
    \centering
    \includegraphics[width=\linewidth]{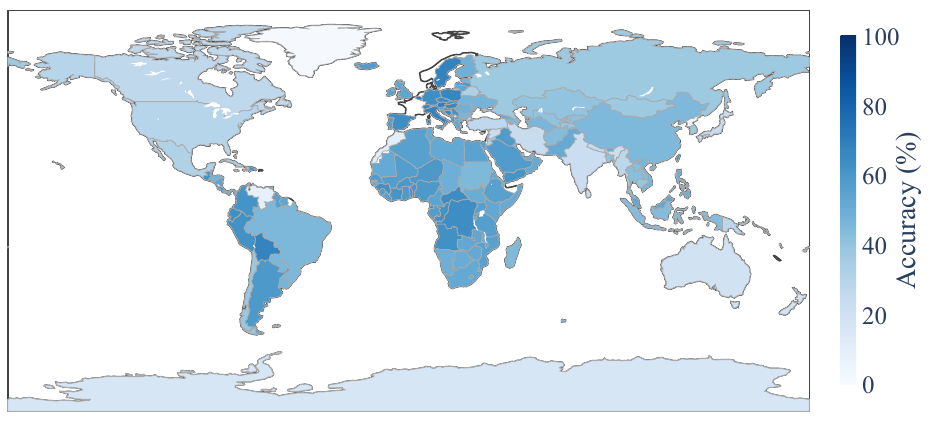}
    \caption{\textbf{Model accuracy aggregated by start country} for the tasks \textsc{TimePlace} and \textsc{TimeTimePlace}. We neither observe biases towards Western nor towards higher-income or population-dense countries.}
    \label{fig:enter-label}
\end{figure}

\clearpage

\subsection{Perplexity Analysis} \label{app:perplexity}

\begin{figure*}[h]
     \centering
     \begin{subfigure}[b]{0.3\textwidth}
         \centering
         \includegraphics[width=\textwidth]{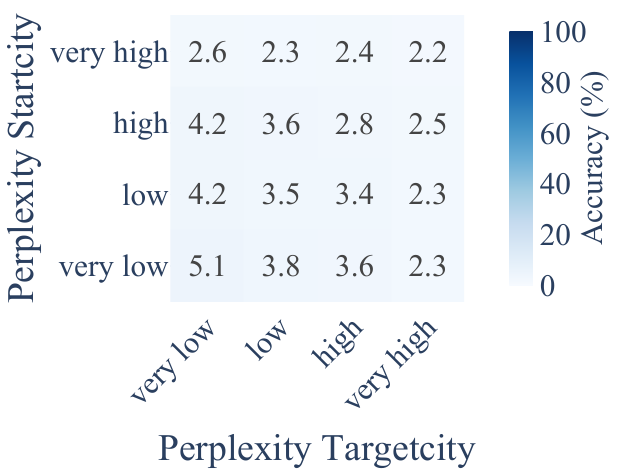}
         \caption{Llama2 7B}
     \end{subfigure}
     \begin{subfigure}[b]{0.3\textwidth}
         \centering
         \includegraphics[width=\textwidth]{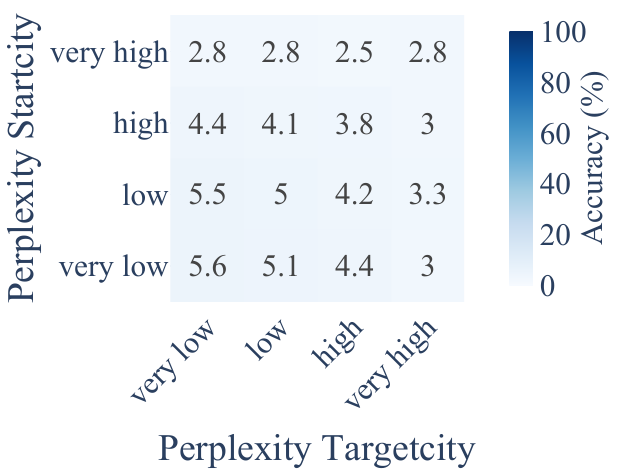}
         \caption{Llama2 13B}
     \end{subfigure}
     \begin{subfigure}[b]{0.3\textwidth}
         \centering
         \includegraphics[width=\textwidth]{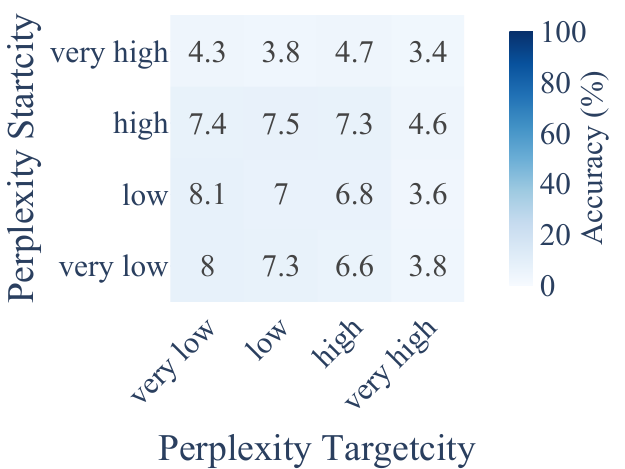}
         \caption{Llama2 70B}
     \end{subfigure}
        \caption{Llama2 accuracy distribution for different City Perplexities}
        \label{fig:ppl_llama2}
\end{figure*}

\begin{figure*}[h]
     \centering
     \begin{subfigure}[b]{0.3\textwidth}
         \centering
         \includegraphics[width=\textwidth]{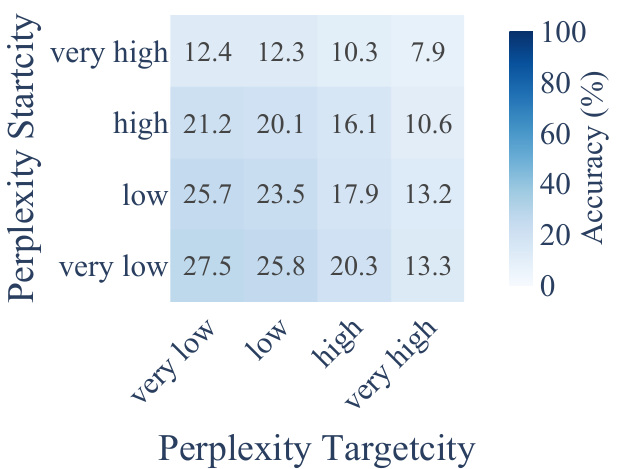}
         \caption{Llama3 8B}
     \end{subfigure}
     \begin{subfigure}[b]{0.3\textwidth}
         \centering
         \includegraphics[width=\textwidth]{img/ppl_l3_l4_Meta-Llama-3-70B-Instruct.pdf}
         \caption{Llama3 70B}
     \end{subfigure}
        \caption{Llama3 accuracy distribution for different City Perplexities}
        \label{fig:ppl_llama3}
\end{figure*}

\begin{figure*}[h]
     \centering
     \begin{subfigure}[b]{0.3\textwidth}
         \centering
         \includegraphics[width=\textwidth]{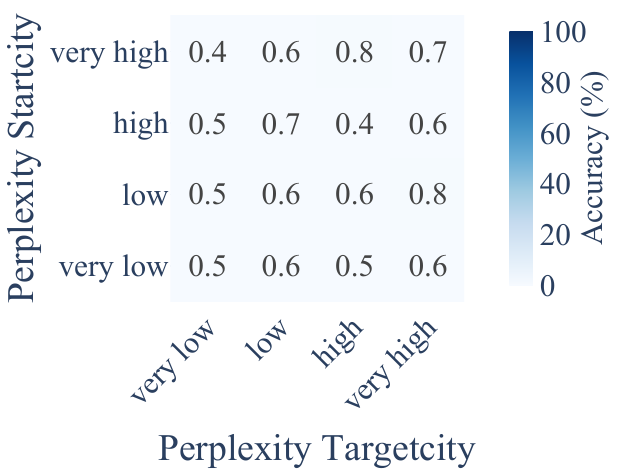}
         \caption{Qwen2 1.5B}
     \end{subfigure}
     \begin{subfigure}[b]{0.3\textwidth}
         \centering
         \includegraphics[width=\textwidth]{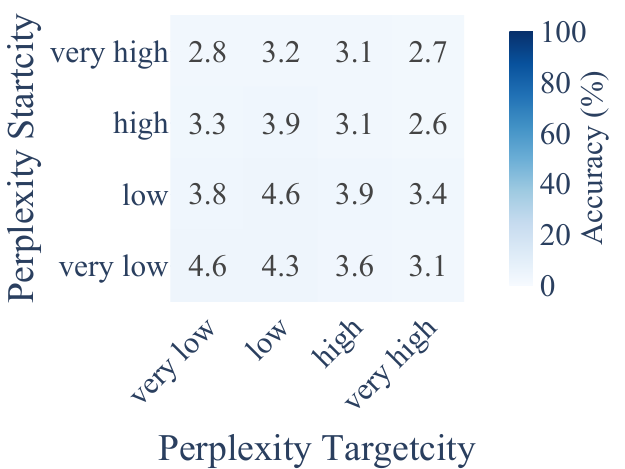}
         \caption{Qwen2 7B}
     \end{subfigure}
     \begin{subfigure}[b]{0.3\textwidth}
         \centering
         \includegraphics[width=\textwidth]{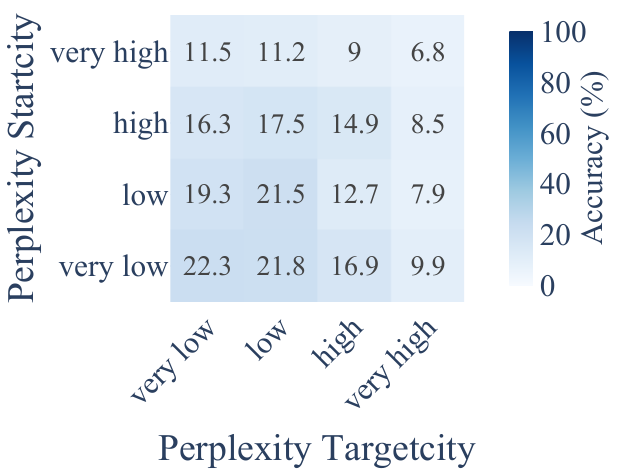}
         \caption{Qwen2 72B}
     \end{subfigure}
        \caption{Qwen2 accuracy distribution for different City Perplexities}
        \label{fig:ppl_qwen2}
\end{figure*}

\clearpage

\section{Qualitative Model Responses}

\paragraph{Qualitative Examples of Model Responses}

We present examples of model responses obtained for the three different instruction types: (1) short, (2) neutral, and (3) CoT, respectively.

\setlength{\tabcolsep}{3pt}
\begin{table}[h]
    \centering
    \small
    \begin{tabular}{p{1cm}|c|p{4cm}|p{1.5cm}|p{5.5cm}|c}
    \toprule
\textbf{Model} & \textbf{Task} & \textbf{Input Prompt} & \textbf{Answer} & \textbf{Model Response} & \textbf{Label} \\
\midrule
Qwen2-1.5B & \textsc{Verification} & Today is Friday, January 6th at 09:28 AM in 2023 in Valletta, Malta. What time is it now in Valletta, Malta? $[...]$ & 2023-01-06 09:28 AM & user 10:47 PM & \xmark \\
Qwen2-7B & \textsc{TimeTimePlace} & Today is Sunday, June 18th at 09:34 PM in 2023 in Mazatlan, Mexico. What time is it in 4 hours in Luxembourg, Luxembourg? $[...]$ & 2023-06-19 10:34 AM & Assistant: Wednesday, June 22nd, 2023, 01:34 AM in Luxembourg, Luxembourg. & \xmark \\
Qwen2-7B & \textsc{TimeTime} & Today is Monday, December 4th at 02:19 AM in 2023 in Nome, United States. What time is it in 7 hours? $[...]$ & 2023-12-04 09:19 AM & answer Tuesday, December 4th at 09:19 AM in 2023 & \cmark \\
\hline
Llama3-8B & \textsc{TimeTimePlace} & Today is Friday, February 3rd at 06:36 AM in 2023 in Gambier Islands, French Polynesia. What time is it in 2 hours in Manila, Philippines? $[...]$ & 2023-02-04 01:36 AM & assistant February 3rd, 2023, 08:36 PM & \xmark \\
Llama3-8B & \textsc{TimeTime} & Today is Tuesday, May 23rd at 05:32 AM in 2023 in Dublin, Ireland. What time is it in 2 hours? $[...]$ & 2023-05-23 07:32 AM & assistant 07:32 AM, May 23rd, 2023 & \cmark \\
Llama3-70B & \textsc{TimePlace} & Today is Thursday, September 14th at 05:24 PM in 2023 in Zagreb, Croatia. What time is it now in Managua, Nicaragua? $[...]$ & 2023-09-14 09:24 AM & assistant September 14th, 2023, 11:24 AM & \xmark \\
Llama3-70B & \textsc{TimeTime} & Today is Sunday, June 4th at 03:27 AM in 2023 in Aden, Yemen. What time is it in 4 hours? $[...]$ & 2023-06-04 07:27 AM & assistant Sunday, June 4th, 07:27 AM & \cmark \\
\hline
Llama2-7B & \textsc{TimeTimePlace} & Today is Saturday, October 21st at 07:47 PM in 2023 in Rothera Point, Adelaide Island. What time is it in 4 hours in Tbilisi, Georgia? $[...]$ & 2023-10-22 06:47 AM & In 4 hours from 07:47 PM in Rothera Point, Adelaide Island on October 21st, 2023, it will be 11:47 PM in Tbilisi, Georgia on October 21st, 2023. & \xmark \\
Llama2-13B & \textsc{Verification} & Today is Saturday, December 2nd at 01:06 AM in 2023 in Astrakhan, Russia. What time is it now in Astrakhan, Russia? $[...]$ & 2023-12-02 01:06 AM & Sure! It is currently Saturday, December 2nd at 01:06 AM in Astrakhan, Russia. & \cmark \\
Llama2-70B & \textsc{TimeTime} & Today is Tuesday, July 18th at 02:09 AM in 2023 in Cocos, Australia. What time is it in 2 hours? $[...]$ & 2023-07-18 04:09 AM & Sure! In 2 hours, it will be: 04:09 AM on Tuesday, July 18th, 2023. & \cmark \\
Llama2-70B & \textsc{TimeTimePlace} & Today is Sunday, February 5th at 12:56 AM in 2023 in Alofi, Niue. What time is it in 7 hours in Brisbane, Australia? $[...]$ & 2023-02-06 04:56 AM & Sure! In 7 hours, it will be 7:56 AM on Sunday, February 5th, 2023 in Brisbane, Australia. & \xmark \\
    \end{tabular}
    \caption{Examples of the model responses on the short instruction template.}
\end{table}

\setlength{\tabcolsep}{2pt}
\begin{table}[h]
    \centering
    \small
    \begin{tabular}{p{1cm}|c|p{3.5cm}|p{1.5cm}|p{7cm}|c}
    \toprule
\textbf{Model} & \textbf{Task} & \textbf{Input Prompt} & \textbf{Answer} & \textbf{Model Response} & \textbf{Label} \\
\midrule
Qwen2-1.5B & \textsc{TimePlace} & Today is Saturday, October 28th at 11:50 PM in 2023 in Vilnius, Lithuania. What time is it now in Chita, Russia? & 2023-10-29 05:50 AM & assistant It is currently 9:50 AM on October 29th, 2023 in Chita, Russia. & \xmark \\
Qwen2-7B & \textsc{TimeTime} & Today is Tuesday, June 27th at 01:36 PM in 2023 in Grand Turk, Turks and Caicos Islands. What time is it in 8 hours? & 2023-06-27 09:36 PM & answer To calculate the time 8 hours from now from 01:36 PM on Tuesday, June 27th, 2023, in Grand Turk, Turks and Caicos Islands, we simply add 8 hours to this time. 01:36 PM + 8 hours = 09:36 PM Therefore, 8 hours from now, it will be 09:36 PM on the same day, Tuesday, June 27th, 2023, in Grand Turk, Turks and Caicos Islands. & \cmark  \\
Qwen2-72B & \textsc{Verification} & Today is Thursday, August 10th at 05:04 PM in 2023 in Novokuznetsk, Russia. What time is it now in Novokuznetsk, Russia? & 2023-08-10 05:04 PM & {\color{red}[Incomplete Generation]} :04 PM on Thursday, August 10th, 2023 in Novokuznetsk, Russia. However, if you're asking for the current time and it's after that date and time, you would need to adjust for the time that has passed since then. Since I can't provide real-time information, you should check the current time using a reliable time source or a device that updates in real-time. & \xmark \\
\hline
Llama3-8B & \textsc{TimeTimePlace} & Today is Friday, August 11th at 08:52 AM in 2023 in Ljubljana, Slovenia. What time is it in 9 hours in Palmer Station, Anvers Island? & 2023-08-11 12:52 PM & assistant A nice geography and time zone question! Ljubljana, Slovenia is in the Central European Summer Time (CEST) zone, which is UTC+2. Palmer Station, Anvers Island is in the Antarctic Time (ANT) zone, which is UTC-3. To find the time in Palmer Station, we need to subtract 5 hours from the current time in Ljubljana. Since it's 08:52 AM in Ljubljana, we subtract 5 hours to get: 08:52 AM - 5 hours = 03:52 AM So, it would be 03:52 AM in Palmer Station, Anvers Island. & \xmark \\
Llama3-8B & \textsc{TimeTime} & Today is Thursday, July 27th at 02:32 PM in 2023 in Brisbane, Australia. What time is it in 7 hours? & 2023-07-27 09:32 PM & assistant Since it's 02:32 PM in Brisbane, Australia, which is in the Australian Eastern Standard Time (AEST) zone, it's already past the standard time zone's 12:00 PM (noon) mark. In 7 hours, it would be: 02:32 PM + 7 hours = 09:32 PM So, in 7 hours, it would be 09:32 PM on Thursday, July 27th, 2023, in Brisbane, Australia. & \cmark \\
Llama3-70B & \textsc{Verification} & Today is Thursday, February 2nd at 12:36 PM in 2023 in Johannesburg, South Africa. What time is it now in Johannesburg, South Africa? & 2023-02-02 12:36 PM & assistant That's an easy one! According to the information you provided, the current time in Johannesburg, South Africa is 12:36 PM on Thursday, February 2nd, 2023. & \cmark \\
\hline
Llama2-7B & \textsc{TimePlace} & Today is Sunday, July 16th at 01:08 PM in 2023 in Broken Hill, Australia. What time is it now in Rarotonga, Cook Islands? & 2023-07-15 05:38 PM & To find out what time it is in Rarotonga, Cook Islands, we need to consider the time difference between Broken Hill, Australia and Rarotonga. Rarotonga is located in the Cook Islands, which are located in the Pacific Ocean, about 3,000 miles (4,800 km) south of Broken Hill, Australia. Rarotonga is 18 hours ahead of Broken Hill. So, if it is 01:08 PM (13:08) on July 16th in Broken Hill, Australia, it would be 6:08 AM on July 16th in Rarotonga, Cook Islands (UTC+13). & \xmark \\
Llama2-13B & \textsc{TimePlace} & Today is Thursday, September 14th at 05:45 PM in 2023 in Tunis, Tunisia. What time is it now in Sydney, Australia? & 2023-09-15 02:45 AM & Sydney, Australia is 10 hours ahead of Tunis, Tunisia. So, if it is 05:45 PM in Tunis on Thursday, September 14th, 2023, it would be 11:45 PM in Sydney on Thursday, September 14th, 2023. & \xmark \\
Llama2-70B & \textsc{TimePlace} & Today is Wednesday, January 11th at 08:58 AM in 2023 in Qostanay, Kazakhstan. What time is it now in Tehran, Iran? & 2023-01-11 06:28 AM & Tehran, Iran is in the Iran Standard Time (IRST) time zone, which is 3 hours and 30 minutes ahead of Qostanay, Kazakhstan. So, if it is 08:58 AM in Qostanay, it would be 12:28 PM in Tehran. & \xmark \\
    \end{tabular}
    \caption{Examples of the model responses on the neutral instruction template.}
\end{table}

\setlength{\tabcolsep}{3pt}
\begin{table}[]
    \centering
    \small
    \begin{tabular}{p{1cm}|c|p{4cm}|p{1.5cm}|p{6cm}|c}
    \toprule
\textbf{Model} & \textbf{Task} & \textbf{Input Prompt} & \textbf{Answer} & \textbf{Model Response} & \textbf{Label} \\
\midrule
%Qwen2-1.5B & \textsc{TimePlace} & Today is Sunday, March 26th at 10:11 PM in 2023 in Tallinn, Estonia. What time is it now in Stockholm, Sweden? Think step by step. & 2023-03-26 09:11 PM & answer To find the current time in Stockholm, Sweden, we need to convert the given time from Tallinn, Estonia into Stockholm's local time. Step 1: Convert the given date and time from Tallinn, Estonia to Stockholm, Sweden. - Date: March 26th, 2023 - Time: 10:11 PM Step 2: Subtract the difference between the two cities' time zones. - The time zone difference between Tallinn (Estonia) and Stockholm (Sweden) is UTC+1. - So, subtracting this difference gives us the local time in Stockholm. Step 3: Calculate the new time in Stockholm. - Since the original time was 10:11 PM on March 26th, 2023, adding the time difference of +1 hour will give us the correct time in Stockholm. Therefore, the current time in Stockholm, Sweden is 11:11 PM on March 27th, 2023. & \xmark \\
Qwen2-7B & \textsc{TimeTime} & Today is Thursday, September 21st at 01:07 PM in 2023 in Caracas, Venezuela. What time is it in 2 hours? Think step by step. & 2023-09-21 03:07 PM & user To find out what time it will be in 2 hours from now in Caracas, Venezuela, given that it's currently Thursday, September 21st at 01:07 PM in 2023, follow these steps: 1. **Add the hours to the current time**: - Current time: 01:07 PM - Adding 2 hours: 01:07 PM + 2 hours = 03:07 PM So, in 2 hours from now, it will be 03:07 PM on Thursday, September 21st, 2023, in Caracas, Venezuela. 2. **Check for any time zone adjustments**: - Caracas uses UTC-4 (or UTC-5 during Daylight Saving Time). However, since we're only adding 2 hours and not considering Daylight Saving Time changes, there's no adjustment needed for this calculation. Therefore, the final answer is that in 2 hours from now, it will be 03:07 PM on Thursday, September 21st, 2023, in Caracas, Venezuela. & \cmark \\
Qwen2-72B & \textsc{TimeTimePlace} & Today is Sunday, January 8th at 03:12 PM in 2023 in Lisbon, Portugal. What time is it in 9 hours in Noronha, Brazil? Think step by step. & 2023-01-08 10:12 PM & Fernando de Noronha, Brazil is 2 hours behind Lisbon, Portugal. If it's 03:12 PM on Sunday, January 8th in Lisbon, then it would be 01:12 PM on the same day in Fernando de Noronha. If you add 9 hours to the current time in Fernando de Noronha: 01:12 PM + 9 hours = 10:12 PM So, in 9 hours from now, it will be 10:12 PM on Sunday, January 8th in Fernando de Noronha, Brazil. & \cmark \\
\hline
%Llama3-8B & \textsc{Verification} & Today is Monday, July 24th at 11:05 AM in 2023 in Tehran, Iran. What time is it now in Tehran, Iran? Think step by step. & 2023-07-24 11:05 AM & assistant Let's break it down step by step: 1. Today is Monday, July 24th, 2023. 2. The current time in Tehran, Iran is 11:05 AM (given). 3. Tehran, Iran is in the Iran Standard Time (IRST) zone, which is UTC+3:30. 4. To find the current time in Tehran, we need to add 3 hours and 30 minutes to the given time. So, let's add 3 hours to 11:05 AM: 11:05 AM + 3 hours = 2:05 PM Then, add 30 minutes: 2:05 PM + 30 minutes = 2:35 PM Therefore, the current time in Tehran, Iran is 2:35 PM. & \xmark \\
Llama3-70B & \textsc{TimeTime} & Today is Thursday, January 26th at 05:10 PM in 2023 in Efate, Vanuatu. What time is it in 6 hours? Think step by step. & 2023-01-26 11:10 PM & assistant Let's break it down step by step. Current time: 05:10 PM (17:10) on Thursday, January 26th, 2023 in Efate, Vanuatu. To find the time in 6 hours, I'll add 6 hours to the current time: 1. Current hour: 17 (05:10 PM) 2. Add 6 hours: 17 + 6 = 23 Since 23 is greater than 24, I'll subtract 24 to get the correct hour: 23 - 24 = -1 So, the hour will be 23 - 1 = 22 (10:00 PM) The minutes remain the same: 10 Therefore, in 6 hours, the time will be 10:10 PM on Thursday, January 26th, 2023 in Efate, Vanuatu. & \xmark \\
\hline
Llama2-7B & \textsc{TimeTime} & Today is Saturday, February 18th at 05:24 AM in 2023 in Port Moresby, Papua New Guinea. What time is it in 1 hours? Think step by step. & 2023-02-18 06:24 AM & Great, let's do this! Current time in Port Moresby, Papua New Guinea: 05:24 AM (18th February 2023) 1 hour from now: 06:24 AM (18th February 2023) So, the time in 1 hour from now is 06:24 AM. & \cmark \\
Llama2-13B & \textsc{TimeTime} & Today is Wednesday, August 2nd at 08:35 AM in 2023 in Manama, Bahrain. What time is it in 8 hours? Think step by step. & 2023-08-02 04:35 PM & To find the time in 8 hours, we need to add 8 hours to the current time in Manama, Bahrain. Current time in Manama, Bahrain: 08:35 AM To add 8 hours: 08:35 AM + 8 hours = 08:35 AM + 8 x 60 = 08:35 AM + 480 = 08:35 AM + 480 = 12:15 PM Therefore, the time in 8 hours from now in Manama, Bahrain is 12:15 PM. & \xmark \\
Llama2-70B & \textsc{Verification} & Today is Friday, August 11th at 11:49 AM in 2023 in Macau, China. What time is it now in Macau, China? Think step by step. & 2023-08-11 11:49 AM & Sure, I can help you with that! 1. First, let's check the time zone of Macau, China. Macau is in the China Standard Time (CST) time zone, which is UTC+8. 2. Next, let's convert the time you provided, 11:49 AM, to military time. In military time, 11:49 AM is written as 1149 hours. 3. Now, let's add the offset for the time zone. Since Macau is in the UTC+8 time zone, we need to add 8 hours to the military time. So, 1149 hours + 8 hours = 1949 hours. 4. Finally, we need to convert the military time back to civilian time. In civilian time, 1949 hours is equivalent to 7:49 PM. Therefore, the current time in Macau, China is 7:49 PM on Friday, August 11th, 2023. & \xmark \\
    \end{tabular}
    \caption{Examples of the model responses on the CoT instruction template.}
\end{table}

\clearpage
\paragraph{Qualitative Analysis Errors Llama3-70B}
We present examples from the human-annotated error analysis of the responses from Llama3-70B for the task \textsc{TimeTimePlace}. 
\setlength{\tabcolsep}{3pt}
\begin{sidewaystable}[]
    \centering
    \small
    \begin{tabular}{p{3.5cm}p{6cm}p{2cm}cccc|ccccccc}
    \toprule
        \textbf{Prompt}  & \textbf{Model Response}   & \textbf{Correct}   &  \textbf{$l_1$}   &  \textbf{$l_1$} &  \textbf{$l_2$} & \textbf{$l_2$}    &  \textbf{$l_1$}   &  \textbf{$l_1$}  &  \textbf{$l_2$}  &  \textbf{$l_2$} & \textbf{Timediff.}  &  \textbf{UTC--Diff.}  \\
         &   & \textbf{Answer}   &  \textbf{UTC}  & \textbf{DST}   & \textbf{UTC}  &  \textbf{DST}    &  \textbf{UTC?}  & \textbf{DST?}   &  \textbf{UTC?}  &  \textbf{DST?} & \textbf{correct? }  &  \textbf{correct? }  \\
         \midrule
         Today is Tuesday, August 8th at 10:52 PM in 2023 in Almaty, Kazakhstan. What time is it in 2 hours in Tallinn, Estonia?  &  Almaty, Kazakhstan is 4 hours ahead of UTC, and Tallinn, Estonia is 3 hours ahead of UTC. So Tallinn is 1 hour behind Almaty. It's currently 10:52 PM in Almaty, so in 2 hours it will be 12:52 AM. In Tallinn, which is 1 hour behind, it will be 11:52 PM."  &  2023-08-08 21:52 &  +6.0  &  +5.0  &  3.0  &  2.0   & \xmark  &  \xmark  &  \cmark  &  \xmark  &  \cmark  & \xmark \\
         \midrule
         Today is Monday, April 10th at 07:06 PM in 2023 in Zagreb, Croatia. What time is it in 1 hours in Lagos, Nigeria?  &  Zagreb, Croatia is in the Central European Summer Time (CEST) zone, which is UTC+2. Lagos, Nigeria is in the West Africa Time (WAT) zone, which is UTC+1. So, Lagos is 1 hour behind Zagreb. Current time in Zagreb: 07:06 PM (19:06) on April 10th, 2023 Add 1 hour to account for the time difference: 08:06 PM (20:06) on April 10th, 2023 So, in 1 hour, it will be 08:06 PM (20:06) on April 10th, 2023 in Lagos, Nigeria."  &  2023-04-10 19:06  &  2.0  &  +1.0  &  1.0    &  1.0 &  \cmark  &  \xmark  &  \cmark  &  \xmark  &  \cmark  &  \xmark \\
         \midrule
        Today is Friday, October 20th at 11:16 PM in 2023 in Vaduz, Liechtenstein. What time is it in 7 hours in Libreville, Gabon?  &  Vaduz, Liechtenstein is in the Central European Time (CET) zone, which is UTC+1. Libreville, Gabon is in the West Africa Time (WAT) zone, which is UTC+1 as well.  So, 7 hours from 11:16 PM on Friday, October 20th, 2023 in Vaduz would be 6:16 AM on Saturday, October 21st, 2023 in Libreville.  &  2023-10-21 05:16   &  +2.0  &  +1.0 &  +1.0   &  +1.0  &  \xmark  &  \cmark  &  \cmark  &  \xmark  &  \cmark  &  \xmark \\
        \midrule
        Today is Friday, January 13th at 12:21 AM in 2023 in Bandar Seri Begawan, Brunei. What time is it in 3 hours in Atka, United States?  &  Bandar Seri Begawan, Brunei is UTC+8. Atka, United States is UTC-10. So Atka is 18 hours behind Bandar Seri Begawan. It is currently 12:21 AM on Friday, January 13th in Bandar Seri Begawan. In 3 hours, it will be 3:21 AM on Friday, January 13th in Bandar Seri Begawan. To find the time in Atka, we need to subtract 18 hours. So it will be 6:21 AM on Thursday, January 12th in Atka.  &  2023-01-12 09:21  &  8.0  &  8.0  &  -10.0  &  -9.0 &     \cmark  &  \xmark  & \cmark  &  \xmark  &  \xmark  &  \xmark \\
\bottomrule
    \end{tabular}
    \caption{Examples of the qualitative error analysis.}
\end{sidewaystable}

\end{document}